\PassOptionsToPackage{dvipsnames,table}{xcolor}
\documentclass[10pt,twocolumn,letterpaper]{article}

\usepackage[pagenumbers]{iccv} % To force page numbers, e.g. for an arXiv version
\usepackage{algorithm}
\usepackage{algorithmic}
\usepackage{tabularx}
\usepackage{amsfonts,bm}
\usepackage{threeparttable}
\usepackage{upgreek}
\usepackage{pifont} % Includes \ding commands
\usepackage{bbding}
\usepackage[normalem]{ulem} % For underlining
\usepackage{makecell} % For better cell alignment
\usepackage{color}
\usepackage{colortbl} % For colored tables
\usepackage{wrapfig}
\usepackage{soul}
\usepackage{wasysym}
\usepackage{xspace}
\usepackage{subcaption}
\usepackage[first=0,last=9]{lcg}
\definecolor{Gray}{gray}{0.85}
\usepackage{multirow}
\usepackage{booktabs} 
\usepackage{tikz}
\usepackage{array}
\usepackage{arydshln}
\usepackage{amsmath}
\usepackage{setspace} % For setting line spacing
\usepackage{microtype}

\usepackage{colortbl}
\definecolor{LightGray}{gray}{0.9}
\definecolor{mygray}{gray}{0.9}

\definecolor{iccvblue}{rgb}{0.21,0.49,0.74}
\usepackage[pagebackref,breaklinks,colorlinks,allcolors=iccvblue]{hyperref}

\title{PathDiff: Histopathology Image Synthesis with Unpaired Text and Mask Conditions}

\author{Mahesh Bhosale \textsuperscript{1}
\and
Abdul Wasi \textsuperscript{1}
\and
Yuanhao Zhai \textsuperscript{1}
\and
Yunjie Tian \textsuperscript{1}
\and
Samuel Border \textsuperscript{2}
\and
Nan Xi \textsuperscript{1}
\and
Pinaki Sarder \textsuperscript{2}
\and
Junsong Yuan \textsuperscript{1}
\and
David Doermann \textsuperscript{1}
\and
Xuan Gong \textsuperscript{3}
\\
  \\
  \textsuperscript{1}University at Buffalo%
  \quad
  \textsuperscript{2}University of Florida%
  \quad
  \textsuperscript{3}Harvard Medical School
}
\begin{document}
\maketitle
\begin{abstract} 
{Diffusion-based generative models have shown promise in synthesizing histopathology images to address data scarcity caused by privacy constraints.
Diagnostic text reports provide high-level semantic descriptions, and masks offer fine-grained spatial structures essential for representing distinct morphological regions. However, public datasets lack paired text and mask data for the same histopathological images, limiting their joint use in image generation. 
This constraint restricts the ability to fully exploit the benefits of combining both modalities for enhanced control over semantics and spatial details.
To overcome this, we propose PathDiff, a diffusion framework that effectively learns from unpaired mask-text data by integrating both modalities into a unified conditioning space. 
PathDiff allows precise control over structural and contextual features, generating high-quality, semantically accurate images. 
PathDiff also improves image fidelity, text-image alignment, and faithfulness, enhancing data augmentation for downstream tasks like nuclei segmentation and classification.
Extensive experiments demonstrate its superiority over existing methods.
Our code is published at \url{https://github.com/bhosalems/PathDiff}.
}
\end{abstract}

\section{Introduction}
\label{sec:intro}
The recent advancements in computational pathology, driven by deep learning, are revolutionizing the field of histopathology by addressing critical challenges in tasks such as nuclei classification and segmentation~\cite{SRINIDHI2021101813}, survival prediction~\cite{Silva2021}, multi-instance learning~\cite{Gadermayr2024}, and transfer learning~\cite{SRINIDHI2021101813}. 
Despite these successes, a significant obstacle persists: the substantial volume of annotated data necessary to effectively train deep learning models. Moreover, the high costs and domain expertise required to annotate such data further aggravate this problem.
Pathologists must start with low magnification to assess tissue architecture and cellular arrangement, then shift to higher magnification to evaluate finer details such as cell morphology, nucleoli appearance, and chromatin density.
Annotating these intricate features is both time-consuming and labor-intensive. For instance, fully annotating the 1k whole slide images from the TCGA dataset\footnote{https://www.cancer.gov/ccg/research/genome-sequencing/tcga} \cite{tcga_brca} would require approximately 40k pathologist hours~\cite{graikos2024learned}. Overcoming these limitations is essential to fully unlock the potential of deep learning in histopathology~\cite{Verghese2023}.

\begin{figure}[t]
        \centering
        \includegraphics[width=\linewidth]{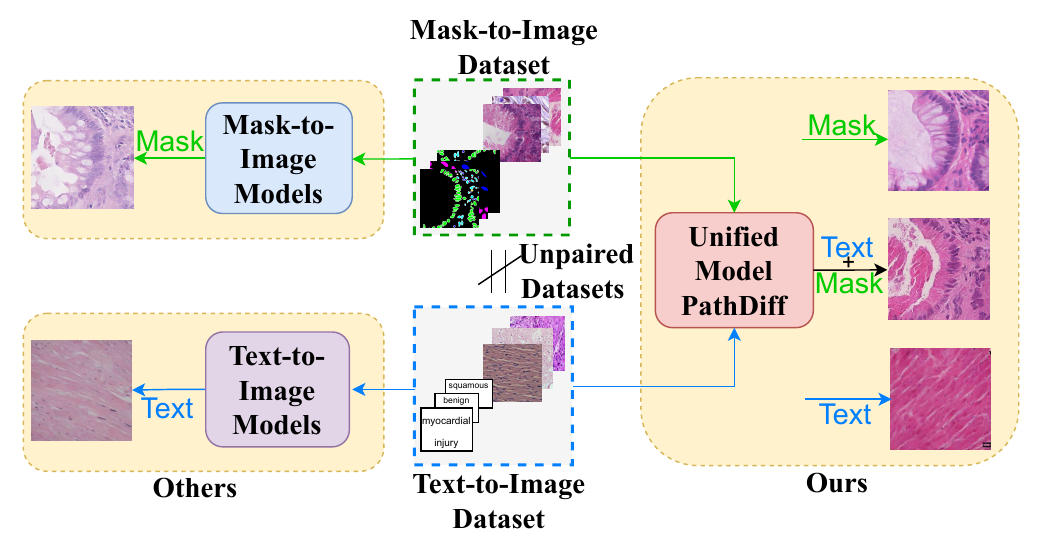}
        \caption{\textbf{PathDiff is trained on unpaired datasets}, integrating two conditional modalities —\textit{Text} and \textit{Mask}— to enable versatile image generation. Unlike unimodal conditional models, PathDiff can generate images conditioned on \textit{Text}, \textit{Mask}, or \textit{both}, allowing greater control and adaptability in image synthesis.}
        \label{fig:PathDiff_Fig1}
\end{figure}

\begin{figure*}[h]
    \centering
    \includegraphics[width=0.92\linewidth]{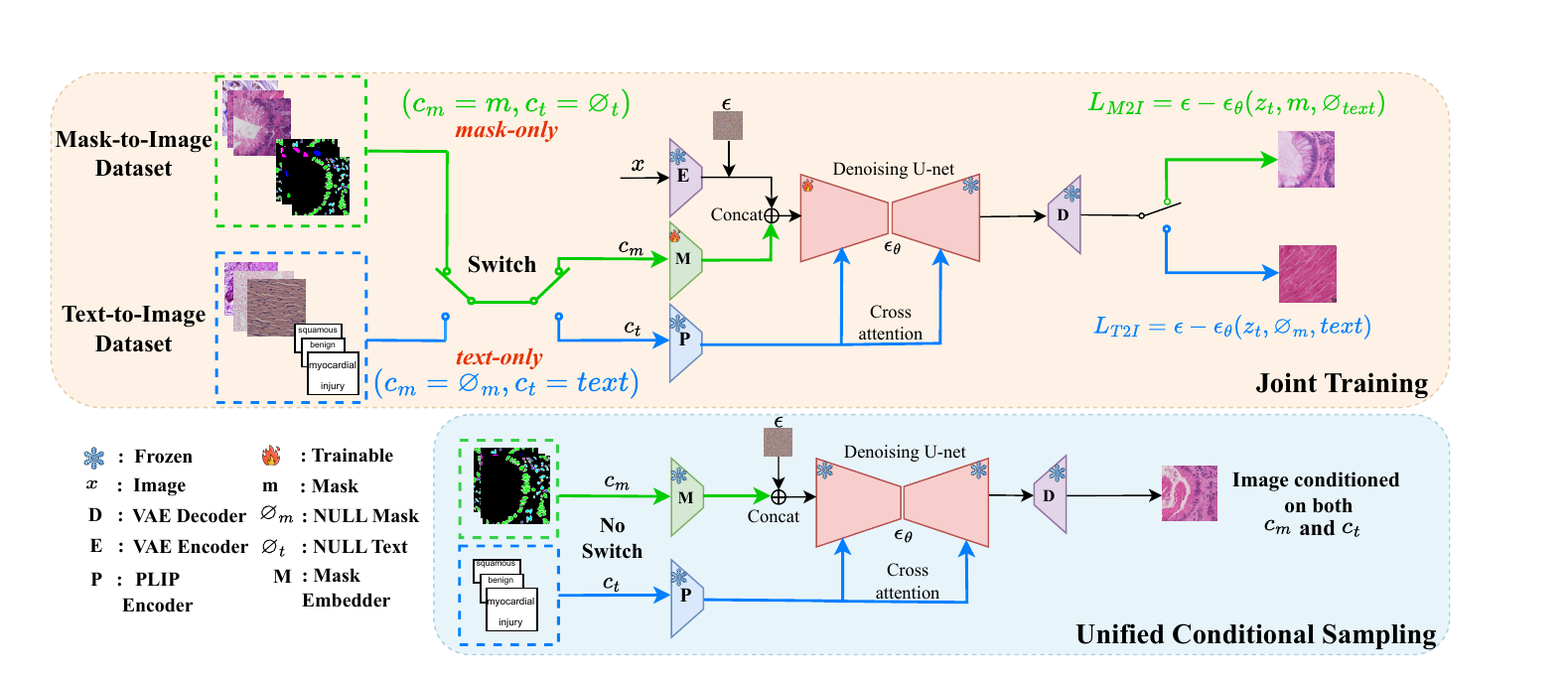}
    \caption{(a) \textbf{PathDiff Training Framework:} A training sample is drawn from either the Mask-to-Image (M2I) or the Text-to-Image (T2I) dataset, determined by the probability $p_{\mathrm{split}}$, which controls the connection of the Switch. The missing condition in each case is set to $\varnothing$. When the Switch selects a sample from the M2I dataset, the Mask-to-Image loss $L_{\mathrm{M2I}}$ is applied; when it selects from the T2I dataset, the Text-to-Image loss $L_{\mathrm{T2I}}$ is used. This approach enables joint training of a single diffusion model on both datasets. (b) \textbf{Image Generation:} During generation, both conditions, $c_m$ (mask) and $c_t$ (text), are applied to produce samples that unify both conditions.}
    \label{fig:PathDiff_Fig2}
\end{figure*}

Generative models in histology \cite{Dolezal2023SyntheticHistology,10030927} 
have thus emerged as a valuable tool to complement existing datasets, extending beyond traditional data augmentation \cite{Levine2020GANPathology, Fernandez2022SASHIMI}. 
More recently, owing to the superior generation quality of diffusion models \cite{NEURIPS2021_49ad23d1, Ho2020Diffusion}, they are widely used for histological image synthesis, either conditioned on diagnostic text reports~\cite{Pathldm, graikos2024learned} or spatial labels such as cell nuclei~\cite{Diffmix, min2024} or regions of interest~\cite{aversa2023diffinfinite}. However, these existing approaches rely solely on a single modality for conditioning, which limits both the quality of control and the amount of data that can be used.

We recognize the importance of considering text and masks, as they provide complementary information. The text offers contextual knowledge, which varies across cancer types, grades, disease stages, or tissue abnormalities, enabling high-level semantic control over the generation process. In contrast, spatial masks capture local structural details, such as cell shapes and types, providing critical spatial information.
Intuitively, simultaneous control of both contextual and spatial information enhances generation. However, no publicly available datasets include open-world text and spatial mask annotations alongside images.

We propose PathDiff, a diffusion framework that unifies text and mask conditions to address the above-mentioned challenges. As shown in \cref{fig:PathDiff_Fig2}, it learns jointly from two independent datasets: one containing image-text pairs and the other containing image-mask pairs. By integrating unpaired text and mask into a single latent space, our model maximizes the use of complementary data and effectively enables generation conditioned on paired text and mask. The main contributions can be summarized as follows:
\begin{itemize}
    \item To address the data scarcity issue in histology image analysis, we propose a novel diffusion-based framework that unifies unpaired text and masks conditions within a single latent space for complementary knowledge exploration.
    \item Our pipeline enables joint learning from independent text-image and mask-image datasets, allowing image generation conditioned on both during inference.
    \item Experiments and qualitative results showcase that our proposed model can effectively integrate the guidance from paired text and mask modalities while the model is trained on unpaired mask and text modalities. 
    \item Owing to the robust and rich joint latent space, empirical evaluations demonstrate that our model outperforms existing approaches across various evaluations, including image fidelity, image-text/mask alignment, and downstream classification and segmentation tasks when conditioned on individual modalities.
  
\end{itemize}

%-------------------------------------------------------------------------
\section{Related Work}
\label{sec: Related work}
Generative adversarial networks~\cite{goodfellow2014generative} have become popular for medical image synthesis~\cite{Butte2023SharpGAN,10.1007/978-3-031-16434-7_39, pmlr-v121-quiros20a}; however, they frequently introduce artifacts, particularly in histopathology images~\cite{Muller-Franzes2023}. In contrast, diffusion-based methods \cite{Ho2020Diffusion} have demonstrated enhanced image quality in both natural and medical images~\cite{Muller-Franzes2023}. Building on this, Classifier Guidance \cite{classifier_guidance} was introduced to condition image generation on specific classes, followed by Classifier-Free Guidance (CFG) \cite{classifier_free}, which eliminates the need for an auxiliary classifier. Latent Diffusion Models (LDMs) \cite{stablediffusion} further enhance the computational efficiency, while ControlNet \cite{controlnet} utilizes CFG to introduce multiple spatial controls in Text-to-Image LDMs \cite{stablediffusion}. Recent approaches for conditional histopathology image generation emphasize text- and mask-based conditioning, which we briefly review in the following sections.

\subsection{Text to Histopathology Image Synthesis}
The application of diffusion models in text-conditioned histopathology image generation remains limited. Summarized reports from the Large Text-Image histopathology dataset TCGA-BRCA~\cite{tcga_brca} were used for text-conditioned image generation in \cite{Pathldm} in conjunction with text annotated Tumor-Infiltrating Lymphocyte (TIL) and Tumor probabilities from off the shelf classifiers. Genome sequencing data from the TCGA-BRCA dataset was used in \cite{10030927}, where heavier weights were assigned to earlier steps and lower weights to later steps in the diffusion process to focus on morphological features. Authors in \cite{graikos2024learned} train diffusion model conditioned on Self Supervised Learnt (SSL) embeddings. An auxiliary diffusion model is trained on SSL embedding paired with Quilt \cite{ikezogwo2023quilt} image embeddings. At the time of inference, text embeddings are used as proxies for image embedding, allowing for text-to-image synthesis. However, note that none of these methods use mask conditions in conjunction with text.

\subsection{Mask to Histopathology Image Synthesis}
The authors of~\cite{aversa2023diffinfinite} propose a hierarchical diffusion model to generate large whole-slide images (WSIs) conditioned on synthesized regions of interest. But, this approach does not incorporate fine-grained, cell-level masks. In~\cite{min2024}, a text-driven approach is used first to generate cell masks, which, along with their distance maps, are then used to condition histopathology image synthesis. However, this method does not allow for fine-grained control over the spatial placement of the masks. In~\cite{yuxinyi_2023}, the diffusion model is trained to synthesize nuclei structures as pixel-level semantic and distance-transform maps, which are then post-processed into instance maps. This is followed by a conditional diffusion model to generate histopathology images. Similarly,~\cite{eccv_min} introduces a cell-point map to synthesize cell-type masks and images jointly. To tackle class imbalance,~\cite{Diffmix} employs a Semantic-Diffusion-Model (SDM)~\cite{SDM} for data synthesis, effectively balancing class variance in nuclei datasets.

In contrast, PathDiff uniquely generates histopathology images by unifying mask and text conditions from unpaired datasets, which is beyond the scope of the works discussed above.
%-------------------------------------------------------------------------
\section{Method}
% We summarize the background in~\cref{subsec:diffusion_background}. In \cref{subsec:pathdiff}, we describe our framework incorporating control from two unpaired datasets.

\subsection{Background}
\label{subsec:diffusion_background}
% \subsubsection{Diffusion models}
% \label{subsubsec:diffusion_model}
\textbf{Diffusion models.}
Diffusion Models~\cite{Ho2020Diffusion} are generative models that gradually add noise to data through a forward diffusion process, followed by a reverse denoising process that reconstructs the original sample.
The forward process corrupts the data sample \( x_0 \) through iterative noise addition controlled by a schedule $\alpha_t$:
%\( \alpha = \{\alpha_t\}_{t=1}^T \):
\begin{equation}
q(x_t | x_{t-1}) = \mathcal{N}(x_t; \sqrt{1 - \alpha_t} \, x_{t-1}, \alpha_t I),
\end{equation}
where $\alpha_t$ is a predefined noise schedule.
The noisy sample $x_t$ can be computed by $x_t =\sqrt{\bar{\alpha}_t} x_0 + \sqrt{1 - \bar{\alpha}_t} \epsilon$, where $\bar{\alpha}_t = \prod_{i=1}^{t} (1 - \alpha_i)$ and $\epsilon \in \mathcal{N}(0,1)$ is a random Gaussian noise.

The reverse process, parameterized by a neural network $p_\theta$, learns a time-conditioned model to remove the noise added at each step.

Latent Diffusion Models (LDMs) work in a compressed latent space \( z_t \) rather than the high-dimensional data space~\cite{stablediffusion}, where the data \( x_0 \) is encoded as \( z_0 \) through an encoder.
LDMs learn to minimize the objective,
\begin{equation}
L = \mathbb{E}_{z_0, t,\epsilon \sim \mathcal{N}(0, 1)} \left[ \| \epsilon - \epsilon_\theta(z_t, t) \|_2^2 \right],
\end{equation}
where
% \( \epsilon \) is random Gaussian noise sampled from \( \mathcal{N}(0, I) \), and 
\( \epsilon_\theta \) is the model’s predicted noise at time \( t \).
Owing to the relevance to score-based generation models, the model estimates the log density of the distribution $z_t$, i.e., $\epsilon_\theta(z_t, t) \approx -\nabla_{z_t} \log p(z_t)$.

% \subsubsection{Guidance}
% \label{subsubsec:diffusion_guidance}
\textbf{Classifier-free guidance.}
In the denoising process, different conditional inputs $c$ (text, image, depth, mask, etc.) can be added to control the generation, so the denoising model predicts $\epsilon_\theta(z_t, t, c)$.
In classifier-free guidance \cite{classifier_free}, a single neural network is used to parameterize both the unconditional denoising diffusion model $p_\theta(z)$ and conditional denoising diffusion model $p_\theta(z|c)$. While training, with some probability $p_{\mathrm{uncond}}$, the unconditional model receives a null token, $\varnothing$ as $c$. $p_{\mathrm{uncond}}$ is set as a hyperparameter. While sampling, a linear combination of conditional and unconditional score estimates is used:
\begin{equation}
    \tilde{\epsilon}_\theta(z_t, t, c) = (1+w)\epsilon_\theta(z_t, t, c) - w \epsilon_\theta(z_t, t)
    \label{Eq:classfier_free_guidance}.
\end{equation}

\subsection{PathDiff}
\label{subsec:pathdiff}
Histopathology image synthesis requires both high-level semantic guidance and precise structural fidelity.
Existing methods~\cite{controlnet} often require paired text and mask data, limiting their practicality due to the scarcity of such paired datasets in the histopathology domain.
To address these challenges, we propose a unified framework, PathDiff, that leverages unpaired text and mask conditions, enabling fine-grained control over semantic and spatial features. This approach eliminates the need for paired data, supporting flexible and realistic image generation for downstream tasks.

% In the denoising diffusion process described in \cref{subsec:diffusion_background}, the condition $c$ and data $x$ are paired within the same dataset $D$, i.e., $(x, c) \in D, \quad \forall (x, c)$, with each $(x, c)$ pair corresponding to one another. In contrast, our PathDIff aims to jointly learn the distribution of unpaired datasets, as shown in \cref{fig:PathDiff_Fig1}.

Formally, let $D_{\mathrm{T2I}} = \{(x_t, c_t)_i\}$ and $D_{\mathrm{M2I}} = \{(x_m, c_m)_i\}$ represent two unpaired datasets, where $D_{\mathrm{T2I}}$ consists of image-text pairs and $D_{\mathrm{M2I}}$ contains image-mask pairs, with no overlapping images between the two datasets.
We aim to learn a latent denoising diffusion model $p_\theta(z_t|t, c_t, c_m)$. At the inference time, $p_\theta$ can generate image samples conditioned on text $c_t$ and mask $c_m$.

\textbf{Joint training on unpaired datasets.}
The training pipeline, illustrated in \cref{fig:PathDiff_Fig2} (a) and detailed in \cref{alg:training}, takes as input triplets of noisy latent, conditional mask, and conditional text, \( (x_t, c_m, c_t) \). We adopt a sampling strategy alternating between T2I and M2I datasets, allowing the model to learn from both sources and effectively integrate text and mask conditions.
Specifically, we jointly train PathDiff by sampling data from the T2I dataset with probability \( p_{\mathrm{split}} \) and from the M2I dataset with probability \( 1 - p_{\mathrm{split}} \).
When sampling from \( D_{\mathrm{T2I}} \) dataset, we set \( c_m = \varnothing_m \), leading to a training triplet of \( (x_t, \varnothing_m, c_t) \).
Similarly, when sampling from \( D_{\mathrm{M2I}} \), we set \( c_t = \varnothing_{t} \), resulting in a training sample of \( (x_m, c_m, \varnothing_t) \).
Following existing methods~\cite{stablediffusion}, we set \( \varnothing_{t} \) as an empty string.
For \( \varnothing_m \), while previous works \cite{SDM,Diffmix,min2024} use a zero vector, we instead use a mask filled with an invalid label, as zero in our dataset indicates a background mask.
Additionally, for robustness, we apply unconditional training by setting both $c_t=\varnothing_{t}$ and $c_m=\varnothing_{m}$ with a small probability $p_{\mathrm{uncond}}$, as suggested in~\cite{controlnet}.
This training pipeline effectively allows PathDiff to learn from both unpaired conditions, enhancing its ability to generate realistic images conditioned on both modalities.

Following ControlNet~\cite{controlnet} to add spatial control over image generation, we duplicate parts of the U-Net’s downsampling and middle layers, adding \textit{zero convolution} layers to these duplicates. Outputs from these layers are integrated into the original U-Net’s skip connections. U-Net was pre-trained on the T2I histopathology dataset before duplicating. The conditional mask is embedded with shallow CNN before being added as input to the U-Net encoder, while text embeddings are extracted from PLIP: pathology CLIP~\cite{PLIP} and cross-attended with U-Net layers. We use VAE from~\cite{Pathldm} trained on the TCGA-BRCA~\cite{tcga_brca} histopathology dataset to encode and decode images. However, as noted by~\cite{Pathldm}, reconstruction loss significantly affects the generation quality; therefore, we study its effect in supplementary. As shown in \cref{fig:PathDiff_Fig2}, all models are frozen except the copied U-Net encoder and shallow mask embedder.

We employ a latent diffusion pipeline~\cite{stablediffusion} to train PathDiff. A shared VAE encoder-decoder is used for both datasets, resulting in a unified latent representation \( z \). This approach assumes that the VAE can compress and reconstruct images from both datasets without significant loss, maintaining a consistent latent representation across domains.

\definecolor{gray}{rgb}{0.5, 0.5, 0.5}

\begin{algorithm}[t]
\footnotesize
\setstretch{1.2}
\caption{\textbf{Joint Training on Unpaired Datasets}}
\label{alg:training}
\begin{algorithmic}[1]
\REQUIRE \textbf{$p_{\text{uncond}}$}: probability of unconditional training, Text-to-Image dataset $\mathbf{D_{T2I}}$, Mask-to-Image dataset $\mathbf{M_{T2I}}$, \textbf{$p_{\text{split}}$}: probability of sampling from $\mathbf{D_{T2I}}$
\STATE \textbf{repeat}

\STATE \hspace{0.1cm} \textbf{if} $\mathbf{u \sim \text{Uniform}[0, 1] \geq p_{\text{split}}}$ \textbf{then} 
\STATE \hspace{0.3cm} $\mathbf{(z_0, c_m=\varnothing_m, c_t) \sim p_1(z_0, c_t)}$ \hfill \textcolor{gray}{\scriptsize $\triangleright$ Sample data from $D_{T2I}$}
\STATE \hspace{0.3cm} $\mathbf{c_t \leftarrow \varnothing_t}$ with probability $\mathbf{p_{\text{uncond}}}$ \hfill \textcolor{gray}{\scriptsize $\triangleright$ Randomly discard text}

\STATE \hspace{0.1cm} \textbf{else}
\STATE \hspace{0.3cm} $\mathbf{(z_0, c_m, c_t=\varnothing_t) \sim p_2(z_0, c_m)}$ \hfill \textcolor{gray}{\scriptsize $\triangleright$ Sample data from $D_{M2I}$}
\STATE \hspace{0.3cm} $\mathbf{c_m \leftarrow \varnothing_m}$ with probability $\mathbf{p_{\text{uncond}}}$ \hfill \textcolor{gray}{\scriptsize $\triangleright$ Randomly discard mask}

\STATE \hspace{0.1cm} $\mathbf{t \sim \text{Uniform}\{1,..,T\}}$ \hfill \textcolor{gray}{\scriptsize $\triangleright$ Sample timestep}

\STATE \hspace{0.1cm} $\mathbf{\epsilon \sim \mathcal{N}(0, I)}$
\hfill \textcolor{gray}{\scriptsize $\triangleright$ Sample noise}
\STATE \hspace{0.1cm} $\mathbf{z_t = \sqrt{\bar{\alpha}_t} \, z_0 + \sqrt{1 - \bar{\alpha}_t} \, \epsilon}$
\hfill \textcolor{gray}{\scriptsize $\triangleright$ Corrupt data}
\STATE \hspace{0.1cm} \textbf{Take gradient step on:} \\
\hspace{0.1cm} $\mathbf{\| \epsilon - \epsilon_\theta(z_t, t, c_m, c_t) \|^2}$ with respect to $\mathbf{\nabla_\theta}$
\hspace{0.2cm} \textcolor{gray}{\scriptsize $\triangleright$ Optimizing PathDiff}
\STATE \textbf{until} converged
\end{algorithmic}
\end{algorithm}

\textbf{Optimization loss.}
PathDiff jointly optimizes the diffusion model $p_{\theta}$ to predict the noise added at every step $t$ of the noising process.
When the training sample is sampled from $D_{\mathrm{T2I}}$ it minimizes loss with respect to sample $x_t \in D_{\mathrm{T2I}}$,
\begin{equation}
L_{\mathrm{T2I}} = \mathbb{E}_{z_0, t,\epsilon \sim \mathcal{N}(0, 1)} \left[ \| \epsilon - \epsilon_\theta(z_t, t, \varnothing_m, c_t) \|_2^2 \right]
\end{equation}
similarly, for training samples from $D_{M2I}$ it minimizes the loss with respect to sample $x_m \in D_{M2I}$,
\begin{equation}
L_{\mathrm{M2I}} = \mathbb{E}_{z_0, t,\epsilon \sim \mathcal{N}(0, 1)} \left[ \| \epsilon - \epsilon_\theta(z_t, t, c_m, \varnothing_t) \|_2^2 \right]
\end{equation}

\begin{figure*}[th!]
    \centering
    \includegraphics[width=\textwidth]{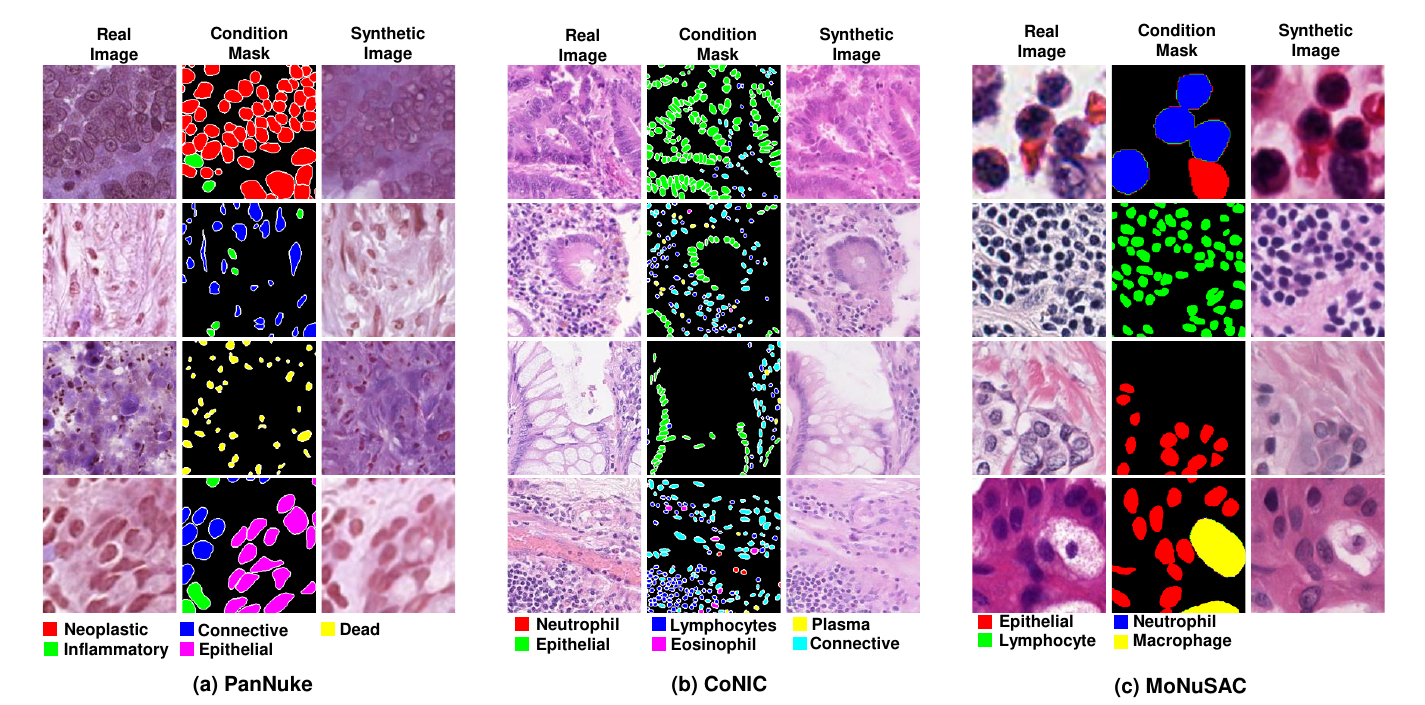}
    \caption{\textbf{Synthetic images generated by PathDiff on Mask to Image datasets} (a) PanNuke \cite{pannuke}, (b) CoNIC \cite{graham2024conic}, (c) MoNuSAC \cite{verma2020multi}. PathDiff closely follows the spatial structures defined by the cell type label map, producing synthetic images that closely resemble the spatial arrangement of real images.}
    \label{fig:qualitative_analysis_m2i}
\end{figure*}

\textbf{Unified conditional sampling.}
Since PathDiff is jointly trained on both \( D_{\mathrm{T2I}} \) and \( D_{\mathrm{M2I}} \) datasets, it can be queried to sample images conditioned on both \( c_t \) and \( c_m \), effectively unifying these conditions as guidance, as illustrated in \cref{fig:qualitative_analysis_t2i_b2i}~(c) and \cref{fig:qualitative_analysis_unified}.
The image generation process is illustrated in \cref{fig:PathDiff_Fig2}~(b). For completeness, we include a sampling algorithm in the supplementary. During inference, we generate images using classifier-free guidance~\cite{classifier_free}, updating the predicted noise with the model using \cref{Eq:classfier_free_guidance}.
The conditioning variable \( c \) can take values from \( \mathbf{c \in \left\{ (\varnothing_m, c_t), (c_m, \varnothing_t), (c_m, c_t) \right\}} \). We can selectively generate images from either distribution by setting one of the conditions to \( \varnothing \).
Specifically,
\[
z_t \sim 
\begin{cases}
    p_{\theta}(z_t | t, c_t) & \text{if } c = (\varnothing_m, c_t), \\
    p_{\theta}(z_t | t, c_m) & \text{if } c = (c_m, \varnothing_t), \\
    p_{\theta}(z_t | t, c_m, c_t) & \text{if } c = (c_m, c_t).
\end{cases}
\]
Images generated according to the conditions are illustrated in \cref{fig:qualitative_analysis_m2i} and \cref{fig:qualitative_analysis_t2i_b2i}~(b) and \cref{fig:qualitative_analysis_unified}. We present extensive experiments on images drawn from each variation of conditions in \cref{sec:experiments}.
% \cref{fig:PCA_ours_star} shows the distribution of two principal components of inception image features against that of real image features from two datasets.
%-------------------------------------------------------------------------

\section{Experiments}
\label{sec:experiments}
We evaluate our method in two main aspects: the quality of the generated images and the effectiveness of using synthetic images as additional training data for downstream task. We evaluate image generation quality in \cref{sec:generation-quality}, focusing on image fidelity, image-mask faithfulness, and alignment quality between images and text. 
Additionally, the utility of the synthetic images in downstream segmentation tasks is detailed in \cref{sec:downstream-tasks}. We also discuss the domain experts' survey results in the supplementary. PathDiff was jointly trained on M2I and T2I datasets; however, we can generate images conditioned on text, mask, or both. Each of these variants is evaluated separately. ControlNet~\cite{controlnet} uses SD \cite{stablediffusion} backbone pre-trained on the T2I dataset for further finetuning on the M2I dataset. Therefore, ControlNet is also evaluated in two variants for a fair and inclusive comparison.

\begin{table*}[h!]
    \centering
    \setlength{\tabcolsep}{0.045cm}
    \begin{tabular}{@{}lccccccccccccccccccccc@{}}
        \toprule
        \multirow{3}{*}{\textbf{Method}} &\multicolumn{1}{c}{\textbf{Condition Modality}} & \multicolumn{4}{c}{\textbf{PanNuke}} & \multicolumn{4}{c}{\textbf{CoNIC}} & \multicolumn{4}{c}{\textbf{MoNuSAC}} \\ \cmidrule(lr){3-6} \cmidrule(lr){7-10} \cmidrule(lr){11-14}
                                         & \multicolumn{1}{c}{} &\multicolumn{2}{c}{\textbf{Train}} & \multicolumn{2}{c}{\textbf{Test}} & \multicolumn{2}{c}{\textbf{Train}} & \multicolumn{2}{c}{\textbf{Test}} & \multicolumn{2}{c}{\textbf{Train}} & \multicolumn{2}{c}{\textbf{Test}} \\ \cmidrule(lr){3-4} \cmidrule(lr){5-6} \cmidrule(lr){7-8} \cmidrule(lr){9-10} \cmidrule(lr){11-12} \cmidrule(lr){13-14} & \textbf{(Train, Sample)}
                                         & \textbf{FID} \(\downarrow\) & \textbf{KID} \(\downarrow\) & \textbf{FID} \(\downarrow\) & \textbf{KID} \(\downarrow\) & \textbf{FID} \(\downarrow\) & \textbf{KID} \(\downarrow\) & \textbf{FID} \(\downarrow\) & \textbf{KID} \(\downarrow\) & \textbf{FID} \(\downarrow\) & \textbf{KID} \(\downarrow\) & \textbf{FID} \(\downarrow\) & \textbf{KID} \(\downarrow\) \\ \midrule
        Diffmix \cite{Diffmix} & (Mask, Mask)           & 7.28            & \underline{0.0492}           & 8.32            & \underline{0.0536}           & 8.09            & 0.0966           & 8.58            & 0.0937           & 9.25            & \underline{0.1036}           & 16.78            & 0.35           \\ 
        SDM \cite{SDM} & (Mask, Mask)                   & \underline{7.14}            & 0.1147           & \textbf{7.13}            & 0.1149           & 8.27            & \underline{0.0709}           & 9.80            & 0.0737          & \underline{7.46}            & 0.1278           & \underline{9.44}            & 0.1340           \\ 
        ControlNet \cite{controlnet} & (Mask+Text, Mask)    & 16.44            & 0.0705           & 15.36           & 0.7722           & \underline{6.76}            & 0.0790           & \underline{6.79}            & \underline{0.0695}           & 20.06            & 0.1050           & 20.56            & \underline{0.1068}           \\ 
        \rowcolor{mygray}
        PathDiff  & (Mask+Text, Mask)                       & \textbf{6.94}            & \textbf{0.0389}           & \underline{7.21}            & \textbf{0.0415}  & \textbf{5.64}            & \textbf{0.0524}           & \textbf{5.54}           & \textbf{0.0488}           & \textbf{6.71}            & \textbf{0.0616}           & \textbf{6.99}     & \textbf{0.0758} 
        % \\
        % Min.et al. \cite{eccv_min} & (Mask, Mask) & 6.50 & 0.0143 & 8.56 & 0.0198 & 30.63 & 0.1645 &  &  & - & - & - & - \\ 
        % \rowcolor{mygray}
        % +PathDiff & (Mask + Text, Mask) & 10.42 & 0.0264 & 12.89 & 0.0321 & - & - & - & - & - & - & - & -
        % Ours-M+T                          & 11.03            & 0.0718           & 12.32            & 0.0952           & \underline{6.28}            & 0.0730           & \ul{6.73}            & 0.0758           & \underline{7.16}            & \ul{0.0782}           & \textbf{6.60}            & \underline{0.0874}           
        \\ \bottomrule
    \end{tabular}
    % }
    \caption {\textbf{Comparison of CLIP-FID and KID across training and test splits} for PanNuke~\cite{pannuke}, CoNIC~\cite{graham2024conic}, and MoNuSAC~\cite{verma2020multi}. PathDiff is trained jointly with T2I dataset: PathCap~\cite{PathCap}. ControlNet~\cite{controlnet} uses SD~\cite{stablediffusion} backbone trained on the PathCap dataset. The best results are in \textbf{bold} and the second best is \underline{underlined}.
    }
    \label{tab:generation_quality_m2i}
\end{table*}

% \begin{table}[!t]
% \footnotesize
% \centering
% \setlength{\tabcolsep}{0.055cm}
% \caption{Ablating $L$ and $s_\mathrm{thr}$ together. Left: LVB, Right: V-MME.}
% \begin{tabular}{ccccccc}
% \toprule
% \backslashbox{$L$}{$s_\mathrm{thr}$} &0.0 & 0.2 & 0.4 & 0.6 & 0.8 & 1.0 \\
% \midrule
% 1 &62.4/63.8 &	62.4/64.0 &	62.5/64.2 &	62.0/64.1 &	61.8/63.8 &	61.9/64.0 \\
% 2 &62.4/63.8 &	62.0/64.0 &	62.4/64.0 &	61.8/63.5 &	61.7/63.4 &	62.0/63.6 \\
% 3 &62.4/63.8 &  \textbf{62.8}/64.0 &	62.6/54.5 &	62.1/64.4 &	62.2/64.4 &	62.1/64.4 \\
% 4 &62.4/63.8 &	62.7/64.1 &	62.7/64.3 &	62.2/64.9 &	62.1/65.0 &	62.2/65.0 \\
% 5 &62.4/63.8 &	62.7/64.1 &	62.2/64.7 &	61.7/65.0 &	61.3/\textbf{65.3} &	61.7/65.2 \\
% 6 &62.4/63.8 &	62.7/64.0 &	62.3/64.5 &	61.8/65.0 &	61.3/65.1 &	61.4/65.1 \\
% \bottomrule
% \end{tabular}
% \label{tab:ablation on depth}
% \vspace{-0.5cm}
% \end{table}

\subsection{Datasets}
\label{sec:datasets}
We use three mask-image histopathology datasets (PanNuke ~\cite{pannuke}, CoNIC~\cite{graham2024conic}, and MoNuSAC~\cite{verma2020multi}) and one text-image dataset (PathCap \cite{PathCap}) to conduct a comparative analysis across multiple metrics, evaluating various diffusion methods and their related downstream tasks.

PanNuke dataset \cite{pannuke} provides annotated histopathology images across 19 tissue types with 189,744 annotated nuclei in five classes (neoplastic, inflammatory, connective, dead, and epithelial). The dataset is highly imbalanced~\cite{CellViT} and one of the most challenging to perform the segmentation task~\cite{ilyas2022tsfd}. %, making it a valuable resource for training and evaluating nuclear segmentation and classification models and supporting model generalization in histopathology analysis.

CoNIC dataset \cite{graham2024conic} % released with the CoNIC 2022 challenge,
is one of the most extensive publicly available histopathology datasets, containing approximately 535,000 labeled colon nuclei across six cell types: epithelial, lymphocytes, plasma cells, eosinophils, neutrophils, and connective tissue cells. 

MoNuSAC \cite{verma2020multi} spans four organs (lung, prostate, kidney, and breast) and includes over 46,000 annotated nuclear boundaries collected from diverse sources for four cell types: epithelial cells, lymphocytes, macrophages, and neutrophils.

PathCap \cite{PathCap} includes 207K pathology image-caption pairs, 197K from PubMed and guidelines, and 10K annotated by expert cytologists in liquid-based cytology (LBC). PathCap has diverse Pan-Cancer image-text pairs spanning multiple organs, making it suitable for joint training. We used a 100K subset of PathCap containing only H\&E-stained histology images to ensure consistency with the three mask-image datasets, which are also H\&E-stained. 

\textbf{Data splits:} All datasets use an 80:20 train/test split, applied consistently to both generation and downstream evaluations. Synthetic examples for downstream task are drawn only from the conditions training split and added during model training to ensure fair comparisons.
% \vspace{-0.25cm}

\subsection{Implementation Details}
\label{sec: Implementation details}
PathDiff is trained on 256\(\times\)256 image patches. We resize patches for the MoNuSAC~\cite{verma2020multi} and PathCap~\cite{PathCap} datasets by taking crops of size 256\(\times\)256 from WSIs. PanNuke~\cite{pannuke} and CoNIC~\cite{graham2024conic} datasets originally have images with size 256\(\times\)256. Our model is trained for 60 epochs on 4 NVIDIA A6000 GPUs with a batch size of 72 and a learning rate of 3.75$\times$10$^{-5}$ with 1000 warmup steps. We use DDIM sampling~\cite{DDIM} with 200 steps and a classifier-free guidance scale set to 1.75 for image generation. We train CellViT on a single NVIDIA RTX A6000 for 130 epochs (with the first 30 epochs frozen), using a batch size of 16 and a learning rate of 0.001, retaining all other training parameters from the original work \cite{CellViT}.

\begin{table}[t!]
    \centering
    % \setlength{\tabcolsep}{5pt}
    % {\fontsize{15}{15}\selectfont
    \resizebox{\linewidth}{!}{%
    \begin{tabular}{lcccccc}
        \toprule
        \multirow{2}{*}{\textbf{Method}} & \multicolumn{2}{c}{\textbf{PanNuke}} & \multicolumn{2}{c}{\textbf{CoNIC}} & \multicolumn{2}{c}{\textbf{MoNuSAC}} \\ \cmidrule(lr){2-3} \cmidrule(lr){4-5} \cmidrule(lr){6-7}
                                         & \textbf{FS1} 
                                         \(\uparrow\) & \textbf{FS2} \(\uparrow\) & \textbf{FS1} \(\uparrow\) & \textbf{FS2} \(\uparrow\) & \textbf{FS1} \(\uparrow\) & \textbf{FS2} \(\uparrow\) \\ \midrule
        
        Diffmix \cite{Diffmix}           & \underline{0.7419}  & \underline{0.6964}           & 0.7597            & 0.7163           & \underline{0.6488}            & \underline{0.6182} \\
        SDM \cite{SDM}                   & 0.7025       & 0.6826 &  0.7433  & 0.7376      & 0.6377  & 0.6104       \\ 
        % \midrule
        ControlNet \cite{controlnet}     & 0.6990       & 0.6821        &   \underline{0.7629}       & \underline{0.7537}     & 0.6317         & 0.6081    \\
        \rowcolor{mygray}
        PathDiff                         & \textbf{0.7437}       & \textbf{0.7406}     & \textbf{0.7873}      & \textbf{0.7632} & \textbf{0.6519}   & \textbf{0.6308} \\ 
        % Min.et.al \cite{eccv_min} & 0.7305 & 0.7032 & - & - & - & - \\ 
        % \rowcolor{mygray}
        % +PathDiff & - & - & - & - & - & -  \\
        % \rowcolor{mygray}
        % Ours-M+T                           & 0.7324       & \ul{0.7187} & 0.7581  & 0.7245      & 0.6412     & \ul{0.6238}        \\ 
        \bottomrule
    \end{tabular}}
    \caption{\textbf{Faithfulness scores} for the PanNuke~\cite{pannuke}, CoNIC~\cite{graham2024conic}, and MoNuSAC~\cite{verma2020multi} datasets. FS1 measures the adherence of synthetic images to the spatial structure of real masks by computing DICE$(M^{pred}_{syn}, M_{real})$. At the same time, FS2 assesses the similarity between synthetic and real images for potential downstream tasks, calculated as DICE$(M^{pred}_{syn}, M^{pred}_{real})$. %Ours-M variant performs best on FS1 and FS2 across all M2I datasets.
    }
    \label{tab:faithfullness_mertric}
\end{table}

\subsection{Image Generation Quality}
\label{sec:generation-quality}

\textbf{Image Generation Fidelity:}
%We evaluate image quality with widely used FID metrics~\cite{FID}. 
To evaluate the quality of generated images, we use CLIP-FID following~\cite{graikos2024learned, parmar2021cleanfid}.
We additionally report Kernel Inception Distance (KID) \cite{kid} to provide an unbiased estimate. As shown in \cref{tab:generation_quality_m2i}, PathDiff achieves the best generation quality in terms of FID and KID compared against the existing mask-to-image generation methods: Diffmix~\cite{Diffmix}, SDM~\cite{SDM} and ControlNet~\cite{controlnet}.
%PathDiff-M achieves the \textbf{best FID and KID} metrics performance across all three M2I datasets, except for the PanNuke test split. 
\cref{fig:qualitative_analysis_m2i} presents qualitative examples generated with the guidance of mask conditions.

\begin{figure*}[ht]
    \centering
    \includegraphics[width=0.9\textwidth]{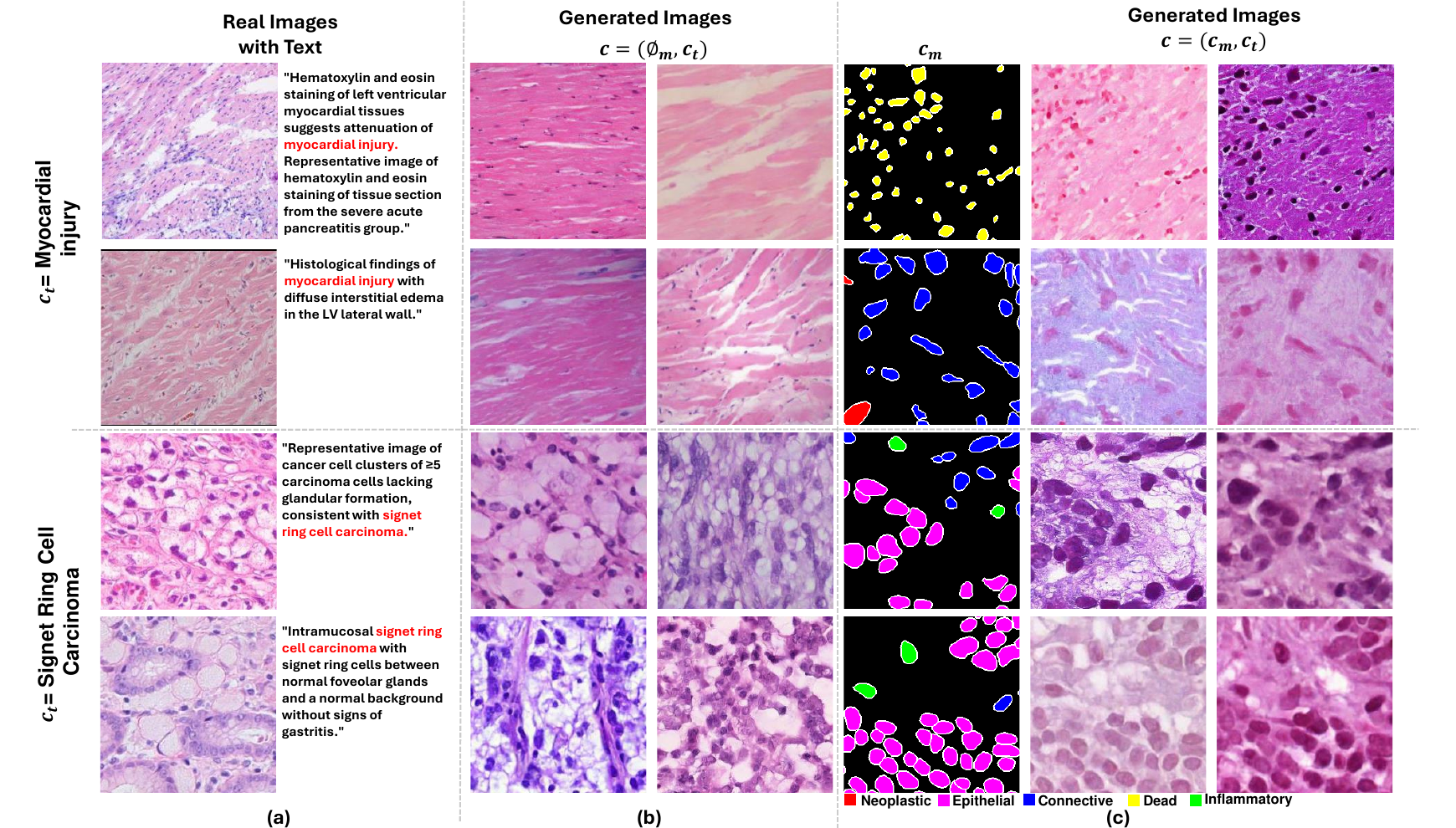}
    \caption{ \textbf{Qualitative examples demonstrate the effectiveness of unifying guidance from random pairings of text and mask conditions.} (a) Real images are shown with their respective text-based descriptions. (b) PathDiff generated images conditioned solely on text replicate the visual attributes observed in real images, with the mask condition set to $\varnothing_m$. (c) PathDiff-generated images conditioned on the same text $c_t$ and an additional mask $c_m$ accurately incorporate the visual features of the real images and adhere to the specified cell mask structure.}
    \label{fig:qualitative_analysis_t2i_b2i}
\end{figure*}

\textbf{Mask-to-Image Faithfulness:}
To evaluate whether the generated images adhere to the spatial guidance provided by the masks, we use the evaluation protocol of SPADE~\cite{park2019SPADE}, also termed as Faithfulness Score (FS) \cite{Kon_AnatomicallyControllable_MICCAI2024}. Auxiliary segmentation model~\cite{CellViT} $S$ is trained on a train split of mask-image datasets. Real images $I_{real}$ from test split $(I_{real}, M_{real})$ are passed through $S$ to get predicted segmentation masks $M^{pred}_{real}$. Synthetic images generated by our model conditioned on masks $M_{real}$ are passed through $S$ to get predicted segmentation masks $M^{pred}_{syn}$. FS1 calculates DICE$(M^{pred}_{syn},M_{real})$, FS2 on other hand calculates DICE$(M^{pred}_{syn}, M^{pred}_{real})$. 
As shown in \cref{tab:faithfullness_mertric}, we attain the best FS1 and FS2 among methods, indicating that the synthetic images closely align with the real masks and reach similarity to real images, enhancing their utility for downstream segmentation tasks. 
Notably, compared with the counterparts \cite{SDM, Diffmix, controlnet},  which also sample from the single mask condition, the joint training strategy demonstrates a superior ability to learn from the limited data.

\textbf{Text-to-Image Alignment:}
\cref{fig:qualitative_analysis_t2i_b2i}(b) presents qualitative samples generated by our model conditioned only on text; images generated replicate visual attributes about the text as observed in real images. \cref{tab:generation_quality_t2i} presents FID and KID scores compared to ControlNet ~\cite{controlnet} conditioned only on text. We also use metric - \textbf{PLIP}~\cite{PLIP} \textbf{cosine similarity} to evaluate the consistency between image and text embeddings. High PLIP image similarity scores indicate better alignment between text and images.
PathDiff has the highest PLIP score of $\geq$ \textbf{24.66}, indicating good alignment between texts and generated images. Our model performs better than ControlNet on FID~\cite{FID} and KID~\cite{kid} across all M2I constituent datasets, indicating large T2I datasets can be jointly trained with a wide range of M2I datasets.

\textbf{Unified Sampling Evaluation:}
No public dataset currently provides (text, mask, image) triplets, which complicates the evaluation of unified conditional sampling. In medical imaging, silver-standard masks~\cite{silver_standard} often replace human (golden-standard) annotations when such labels are unavailable. We use DeepCMorph~\cite{labelpred}—an encoder-decoder model extensively trained on heterogeneous data—to generate silver-standard masks for PathCap. These masks, combined with PathCap text, form triplets for inference and evaluation. Because CONIC~\cite{graham2024conic} covers all nuclei types targeted by DeepCMorph, we employ its checkpoints for these experiments. To check whether the masks generated by DeepCMorph are reasonable, we randomly pair the text in PathCap with masks in the CONIC dataset and use this pairing as conditions for a generation, as indicated by PathDiff-R.
\begin{table*}[th!]
    \centering
    \setlength{\tabcolsep}{0.018cm}
    {\fontsize{7}{7}\selectfont
    % \resizebox{\linewidth}{!}{%
    \resizebox{\textwidth}{!}{

    \begin{tabular}{@{}lcccccccccccccccccccccccc@{}}
        \toprule
        \multirow{3}{*}{\textbf{Method}} & \multicolumn{1}{c}{\textbf{Condition Modality}} & \multicolumn{6}{c}{\textbf{M2I Dataset: PanNuke}} & \multicolumn{6}{c}{\textbf{M2I Dataset: CoNIC}} & \multicolumn{6}{c}{\textbf{M2I Dataset: MoNuSAC}} \\ \cmidrule(lr){3-8} \cmidrule(lr){9-14} \cmidrule(lr){15-20}
                                         & \multicolumn{1}{c}{\textbf{}} & \multicolumn{3}{c}{\textbf{Train}} & \multicolumn{3}{c}{\textbf{Test}} & \multicolumn{3}{c}{\textbf{Train}} & \multicolumn{3}{c}{\textbf{Test}} & \multicolumn{3}{c}{\textbf{Train}} & \multicolumn{3}{c}{\textbf{Test}} \\ \cmidrule(lr){3-5} \cmidrule(lr){6-8} \cmidrule(lr){9-11} \cmidrule(lr){12-14} \cmidrule(lr){15-17} \cmidrule(lr){18-20}
                                         & \textbf{(Train,Sample)} & \textbf{FID} \(\downarrow\) & \textbf{KID} \(\downarrow\) & \textbf{PLIP} \(\uparrow\)  & \textbf{FID} \(\downarrow\) & \textbf{KID} \(\downarrow\) & \textbf{PLIP} \(\uparrow\)  & \textbf{FID} \(\downarrow\) & \textbf{KID} \(\downarrow\) & \textbf{PLIP} \(\uparrow\)  & \textbf{FID} \(\downarrow\) & \textbf{KID} \(\downarrow\) & \textbf{PLIP} \(\uparrow\)  & \textbf{FID} \(\downarrow\) & \textbf{KID} \(\downarrow\) & \textbf{PLIP} \(\uparrow\)  & \textbf{FID} \(\downarrow\) & \textbf{KID} \(\downarrow\) & \textbf{PLIP} \(\uparrow\)  \\ \midrule
        ControlNet~\cite{controlnet}  & (Mask+Text,Text)   & 45.39        & 0.0872        & 23.27 & 45.04       & 0.0886        & 22.95 & 26.32        & 0.1275        & 21.97 & 25.10        & 0.1269        & 21.98 & 28.66       & 0.1111       & 21.88 & 29.86       & 0.1091       & 21.76 \\
        \rowcolor{mygray}
        PathDiff  & (Mask+Text,Text) & \textbf{18.52}        & \textbf{0.0619}        & \textbf{24.18} &  \textbf{19.60}       & \textbf{0.0644}        & \textbf{24.05} & \textbf{18.97}        & \textbf{0.0640}       & \textbf{24.03} &    \textbf{19.96}     & \textbf{0.0655}       & \textbf{24.17} & \textbf{13.72}          & \textbf{0.0538}         & \textbf{24.17} & \textbf{14.04}            & \textbf{0.0557}           & \textbf{24.66} \\

        \bottomrule
    \end{tabular}}
    \caption{\textbf{Comparison of CLIP-FID, KID, and PLIP similarity scores} for training and test splits for text-image dataset: PathCap. PathDiff is jointly trained with three M2I datasets: PanNuke~\cite{pannuke}, CoNIC~\cite{graham2024conic}, and MoNuSAC~\cite{verma2020multi}. PLIP similarity scores on real PathCap~\cite{PathCap} train and test splits are \textbf{26.34} and \textbf{26.56}, respectively, as a reference for the comparison.}
    \label{tab:generation_quality_t2i}}
\end{table*}

\begin{figure*}[ht]
    \centering
    \includegraphics[width=0.9\textwidth]{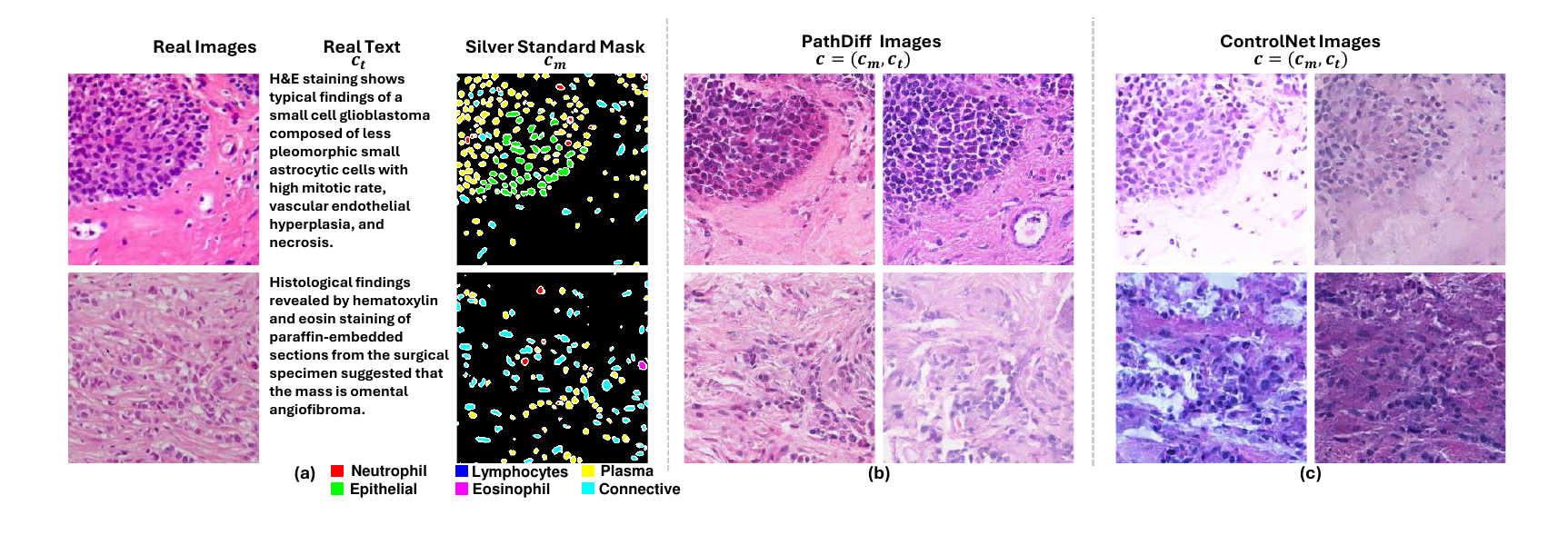}
    \caption{ \textbf{Qualitative examples demonstrate the effectiveness of unifying guidance from paired text and silver standard masks} generated by \cite{labelpred} from PathCap dataset~\cite{PathCap} (a) Real images and Text with silver masks (b) PathDiff generated images (c) ControlNet generated images.}
    \label{fig:qualitative_analysis_unified}
\end{figure*}
In \cref{tab:label_pred}, PathDiff and ControlNet rely on silver-standard masks from PathCap. Even with random pairing, PathDiff-R surpasses ControlNet in FID while maintaining a similar PLIP score. PathDiff attains the best FID and PLIP results, indicating a meaningful pairing of silver-standard masks with text. Furthermore, comparing PathDiff’s FID on CONIC in \cref{tab:generation_quality_t2i} to the metrics in \cref{tab:label_pred} shows that a joint conditioning space yields richer images with improved FID values while preserving text-image similarity, as reflected by a comparable PLIP score. The approach still yields a performance boost even with these silver-standard masks. Note that the models have never seen silver standard masks of the PathCap dataset in the training process. \cref{fig:qualitative_analysis_unified} shows faithful image generation with silver standard masks and texts.
\begin{table}[t!]
    \centering
    \setlength{\tabcolsep}{0.040cm}
    \resizebox{\columnwidth}{!}{ 
    \begin{tabular}{lcccc}
        \toprule
        \textbf{Data Domain} & \textbf{Condition Modality} & \textbf{FID} \(\downarrow\) & \textbf{KID} \(\downarrow\) & \textbf{PLIP} \(\uparrow\) \\ 
        \midrule 
        ControlNet & (Mask+Text, Mask+Text) & 15.72 & \textbf{0.0699} & 22.03 \\
        \rowcolor{mygray}
        PathDiff-R   & (Mask+Text, Mask+Text) & 11.12 & 0.1059 & 22.09 \\
        \rowcolor{mygray}
        PathDiff   & (Mask+Text, Mask+Text) & \textbf{10.54} & 0.0766 & \textbf{24.02} \\
        \bottomrule 
    \end{tabular} }
    \caption{\textbf{Comparison of CLIP-FID, KID, and PLIP scores using text-silver standard mask pairs versus random text-mask pairs}. Text is drawn from PathCap \cite{PathCap}, with masks predicted by DeepCMorph \cite{labelpred}. Random pairings use masks from CONIC \cite{graham2024conic} and texts from PathCap. Both ControlNet \cite{controlnet} and PathDiff rely on silver standard masks, whereas PathDiff-R uses random pairings.}
    \label{tab:label_pred}
\end{table}

\begin{table*}[h!]
    \centering
    % {\fontsize{11}{11}\selectfon
    \resizebox{0.95\linewidth}{!}{%
    \begin{tabular}{llcccccccccc}
        \toprule
        \multirow{2}{*}{\textbf{Dataset}} & \multirow{2}{*}{\textbf{Method}} & \multirow{2}{*}{\textbf{Train Data}} & \multicolumn{4}{c}{\textbf{Segmentation}} & \multicolumn{3}{c}{\textbf{Classification}} \\
        \cmidrule(lr){4-7} \cmidrule(lr){8-10}
        & & & \textbf{Dice} \, \(\uparrow\) & \textbf{Jaccard} \, \(\uparrow\) & \textbf{AJI} \, \(\uparrow\) & \textbf{HD (95)} \, \(\downarrow\) & \textbf{F1} \(\uparrow\) &
        \textbf{Precision} \, \(\uparrow\) & \textbf{Recall} \, \(\uparrow\) \\
        \midrule
        \multirow{5}{*}{PanNuke} 
         & Baseline & Real & 0.7834 &0.6951 & 0.6904 & 6.81 & 0.7826 & 0.7472 & \underline{0.8215} \\
         \cmidrule(lr){2-10}
        & w/ SDM & Real+Synth & 0.7891 & 0.6816 & 0.6729 & 6.26 & 0.7792 & \underline{0.8014} & 0.7581 \\
        
        & w/ ControlNet~\cite{controlnet} & Real+Synth & \underline{0.8093} & \underline{0.6979} & \underline{0.6954} & \underline{5.87} & \underline{0.8134} & \textbf{0.8354} & 0.7925 \\
        % & w/ Diffmix & \textbf{0.8210} & \ul{0.7005} & \ul{0.6987} & \textbf{4.28} & \textbf{0.8230} & \ul{0.8179} & \ul{0.8466} \\
        % \cmidrule(lr){2-9
        
        \rowcolor{mygray}
        & w/ PathDiff & Synth & 0.7311 & 0.6538 & 0.6539 & 7.73 & 0.7140 & 0.7039 & 0.7244 \\
        \rowcolor{mygray}
        & w/ PathDiff & Real+Synth & \textbf{0.8164} & \textbf{0.7122} & \textbf{0.7117} & \textbf{5.12} & \textbf{0.8161} & 0.7827 & \textbf{0.8524} \\
        \midrule

        % \rowcolor{mygray}
          % & w/ Ours-M+T & 0.7947 & \ul{0.7040} & \ul{0.7018} & \ul{5.74} & 0.7949 & 0.7803 & \ul{0.8100} \\
         
        % & w/ Min.et.al\cite{eccv_min} & Real+Synth  & 0.8112 & 0.7187 & 0.7187 & 5.03 & 0.83 & 0.8217 & 0.8403 \\
        % \rowcolor{mygray}
        % & w/ Min.et.al+PathDiff & Real+Synth & - & - & - & - & - & - & - \\
        % \midrule
        \multirow{5}{*}{CoNIC} 
       & Baseline & Real & 0.7927 & 0.6978 & 0.6917 & 4.38 & 0.7614 & 0.7153 & 0.8138 \\ \cmidrule(lr){2-10}
        & w/ SDM & Real+Synth & 0.8007 & 0.6846 & 0.6846 & 4.50 & 0.7883 & 0.7620 & \underline{0.8164} \\
        
        & w/ ControlNet~\cite{controlnet} & Real+Synth & \underline{0.8076} & \underline{0.6999} & \underline{0.6950} & \underline{3.16} & \underline{0.8000} & \textbf{0.8344} & 0.7683 \\
        % & w/ Diffmix & \ul{0.8219} & \ul{0.7176} & \textbf{0.7163} & 3.84 & 0.7895 & 0.7526 & \ul{0.8302} \\ 
        % \cmidrule(lr){2-9}
        \rowcolor{mygray}
        & w/ PathDiff & Synth & 0.7612 & 0.6480 & 0.6473 & 6.37 & 0.7449 & 0.7783 & 0.8107 \\
        \rowcolor{mygray}
        & w/ PathDiff & Real+Synth & \textbf{0.8356} & \textbf{0.7195} & \textbf{0.7141} & \textbf{2.97} & \textbf{0.8053} & \underline{0.7790} & \textbf{0.8328} \\
        % \rowcolor{mygray}
          % & w/ Ours-M+T & \ul{0.8114} & \ul{0.7003} & \ul{0.7003} & 3.41 & 0.7751 & 0.7572 & 0.7938 \\
          
        % & w/ Min.et.al\cite{eccv_min} & Real+Synth & - & - & - & - & - & - & - \\
        % \rowcolor{mygray}
        % & w/ Min.et.al+PathDiff & Real+Synth & - & - & - & - & - & - & - \\
        \midrule
        \multirow{5}{*}{MoNuSAC} 
        & Baseline & Real & \underline{0.7089} & \underline{0.6131} & 
        \underline{0.6097} & 6.77 & 0.6652 & 0.6383 & 0.6944 \\ \cmidrule(lr){2-10}
        & w/ SDM & Real+Synth &0.6783 & 0.5744 & 0.5731 & \underline{6.49} & 0.6685 & \underline{0.6940} & 0.6448 \\
       
        & w/ ControlNet~\cite{controlnet} & Real + Synth & 0.6940 & 0.5852 & 0.5800 & 6.54 & 0.6843 & 0.6533 & 0.7183 \\
        % & w/ Diffmix & \textbf{0.7385} & 0.6079 & 0.6026 & \textbf{5.17} & \ul{0.7096} & \textbf{0.7323} & 0.6882 \\ 
        % \cmidrule(lr){2-9}
        \rowcolor{mygray}
        & w/ PathDiff & Synth & 0.6671 & 0.5894 & 0.5894 & 7.07 & \underline{0.6851} & 0.6345 & \textbf{0.7445}  \\
        \rowcolor{mygray}
        & w/ PathDiff & Real+Synth & 
        \textbf{0.7221} & \textbf{0.6197} & \textbf{0.6192} & \textbf{5.88} & \textbf{0.7115} & \textbf{0.6951} & \underline{0.7286} \\
        % \rowcolor{mygray}
          % & w/ Ours-M+T & \ul{0.7124} & 0.6072 & 0.6067 & \ul{6.25} & \ul{0.6861} & 0.6569 & \ul{0.7187} \\
         
        % & w/ Min.et.al\cite{eccv_min} & Real+Synth & - & - & - & - & - & - & - \\
        % \rowcolor{mygray}
        % & w/ Min.et.al+PathDiff & Real+Synth &- & - & - & - & - & - & - \\
        \bottomrule
    \end{tabular}}
     \caption{\textbf{Comparison of segmentation and classification metrics} across different datasets and methods on the CellViT model. The Baseline is trained with real data, and the rest of the methods have synthetic data added in equal proportion to the train set. }  
    \label{tab:ds_table}
\end{table*}

\subsection{Downstream tasks}
To evaluate the utility of the generated images, we use them as additional training data for the segmentation model. CellViT \cite{horst2024cellvit} (SAM-B variant) backbone with HoVer-Net \cite{graham2019hover} as decoder. \cref{tab:ds_table} compares segmentation and classification performance in multiple metrics, including Dice, Jaccard, AJI (Aggregated Jaccard Index), HD95~\cite{huttenlocher1993comparing} (95th percentile Hausdorff Distance), F1 score, precision and recall.
We note that our proposed method consistently improves the performance across various metrics on all three datasets.
In segmentation tasks on PanNuke, PathDiff trained on real and synthetic data achieves the highest Dice score (0.8164) and performs well across other metrics like Jaccard (0.7122) and AJI (0.7117). Similarly, in classification tasks on PanNuke, PathDiff achieves a strong F1 score of 0.8161 with balanced precision and recall. On the CoNIC dataset, PathDiff again leads in segmentation with a Dice score of 0.8356 and performs well in classification with an F1 score of 0.8053. On MoNuSAC, where performance is generally lower across all methods, PathDiff still outperforms others in segmentation with a Dice score of 0.7221 and achieves an F1 score of 0.7115 for classification. Importantly, training solely with synthetic images still achieves performance comparable to the baseline. This outcome suggests that the synthetic images are robust and simulate the original data effectively. It is pertinent to mention that when evaluating the classification and segmentation performance, the split used included only real test data.
\label{sec:downstream-tasks}
\section{Conclusion}
To address data scarcity in histopathology image analysis, we propose a novel diffusion framework that simultaneously leverages diagnostic reports as contextual guidance and mask inputs for precise spatial control. Incorporating these two components ensures more accurate and meaningful image generation, especially when high-quality annotated data is limited. We show that a joint conditional diffusion model can unify conditions from unpaired data. At inference, classifier-free guidance generates images conditioned on both modalities, addressing data scarcity challenges. This approach optimizes the use of existing data, achieving superior results in both image quality and downstream tasks like segmentation and classification.
{
    \small
    \bibliographystyle{ieeenat_fullname}
    \bibliography{main}
}

\maketitlesupplementary

\section{Scaling Augmentation in Downstream Tasks}

To evaluate the impact of synthetic data augmentation on downstream tasks, we designed three augmented sets for training the CellViT~\cite{CellViT} model on the PanNuke~\cite{pannuke} dataset. These augmented datasets were generated from the training split, conditioned on mask, and evaluated on the real test split. They were incrementally added to the training process while keeping the size of the real train split constant.

\begin{figure*}[t]
    \centering
    \begin{subfigure}[b]{0.25\textwidth}
        \includegraphics[width=\textwidth]{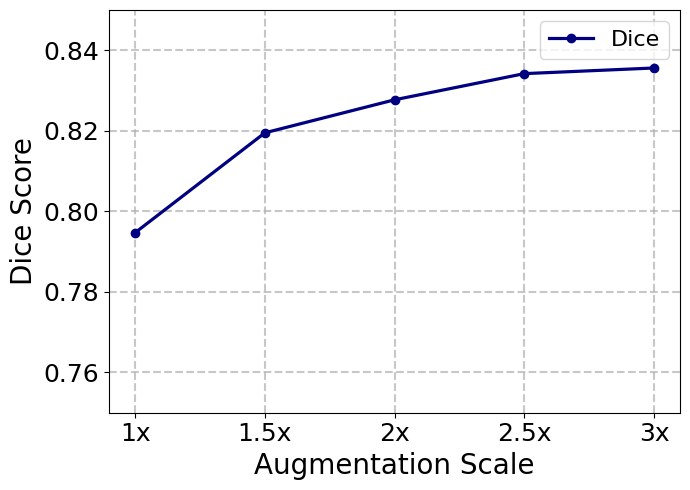}
        \caption{Segmentation - Dice}
        \label{fig:image1}
    \end{subfigure}
    \hfill
    \begin{subfigure}[b]{0.24\textwidth}
        \includegraphics[width=\textwidth]{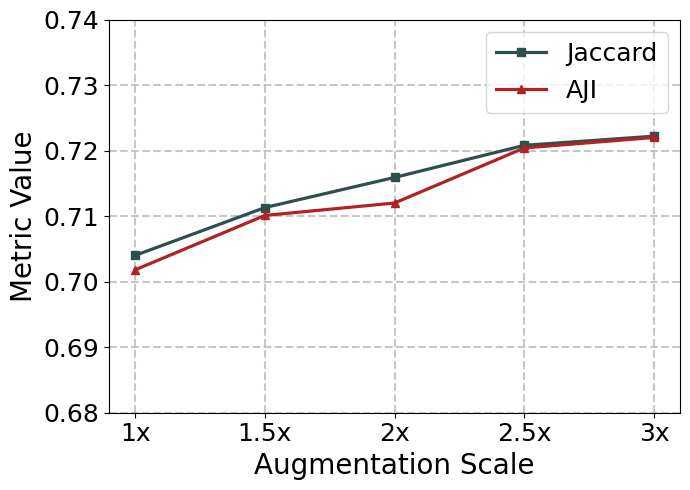}
        \caption{Segmentation - Jaccard \& AJI}
        \label{fig:image2}
    \end{subfigure}
    \hfill
    \begin{subfigure}[b]{0.24\textwidth}
        \includegraphics[width=\textwidth]{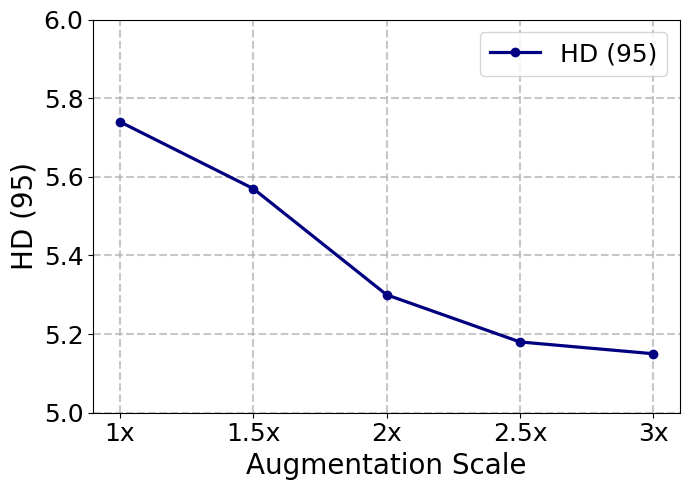}
        \caption{Segmentation - HD (95)}
        \label{fig:image3}
    \end{subfigure}
    \hfill
    \begin{subfigure}[b]{0.24\textwidth}
        \includegraphics[width=\textwidth]{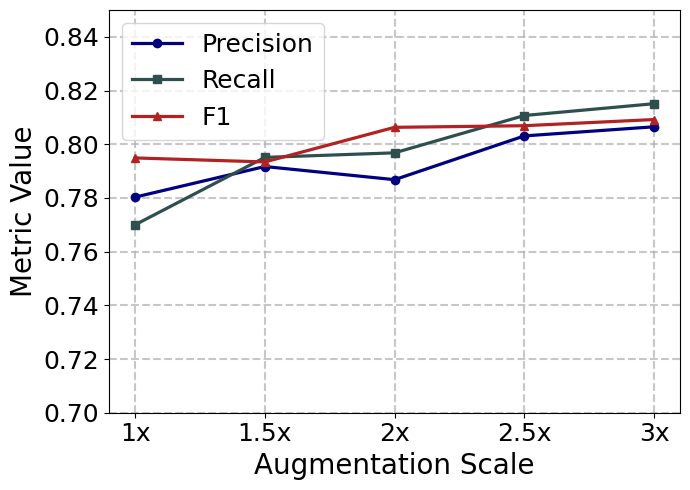}
        \caption{Classification}
        \label{fig:image4}
    \end{subfigure}
    \caption{\textbf{Comparison of segmentation and classification metrics on the PanNuke~\cite{pannuke} dataset across augmentation scaling factors.} The addition of PathDIff-generated synthetic data consistently increases downstream classification and segmentation performance. 1x uses one \textit{synthetic} augmentation set equal to the \textit{real} train split size; 1.5x adds another \textit{synthetic} set equal to 1.5 times the \textit{real} train split size and so on. After 2.5x, the performance metrics plateau.}
    \label{fig:cs_plots}
\end{figure*}

% Augmentation levels are defined as: 1x corresponds to a \textit{synthetic} set equal in size to the real train split, while 1.5x includes a synthetic set 1.5 times the size of the real train split, used alongside the real data for training, and so on. 

Scaling the augmentation set progressively improves the classification and segmentation performance of the training data, as shown in \cref{fig:cs_plots}, with the 3x and 2.5x augmented synthetic datasets outperforming the relatively smaller ones on all metrics. After 2.5x, the performance metrics plateau. These results demonstrate that PathDiff effectively contributes valuable synthetic data in every augmented set, consistently improving model performance across all downstream tasks as the size of the synthetic set increases, highlighting PathDiff generates diverse high-quality data for histopathology image analysis.

\begin{figure*}[t]
        \centering        \includegraphics[width=0.78\linewidth]{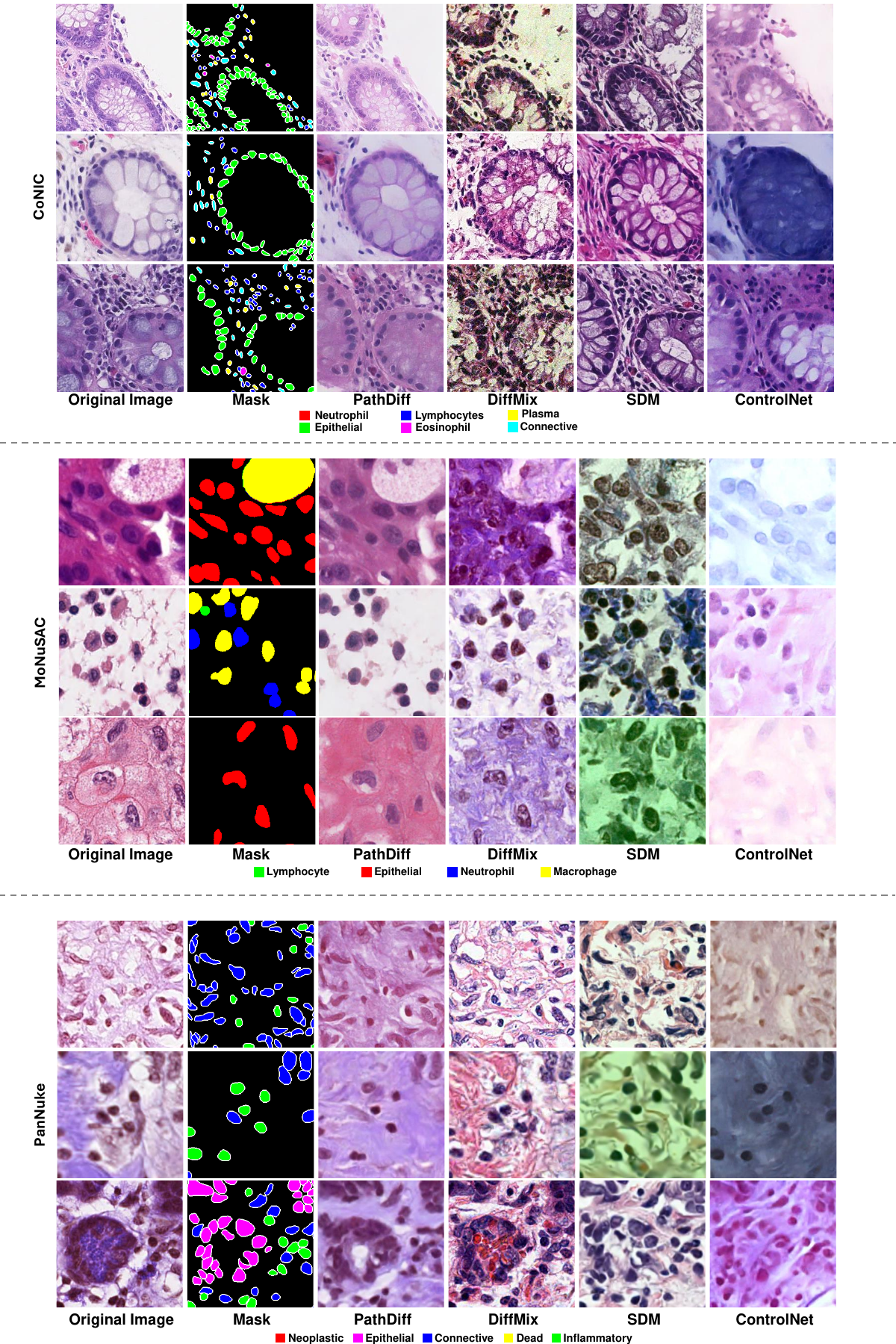}
        \caption{\textbf{Qualitative comparison} of synthetic images generated by PathDiff,  DiffMix~\cite{Diffmix}, SDM~\cite{SDM}, and ControlNet~\cite{controlnet} on the CoNIC~\cite{graham2024conic}, MoNuSAC~\cite{verma2020multi}, and PanNuke~\cite{pannuke} datasets.}
        \label{fig:qual_combined}
\end{figure*}

\section{Qualitative Comparison of Synthetic Images}
In this section, we present a qualitative comparison of synthesized images generated by PathDiff, DiffMix~\cite{Diffmix}, SDM~\cite{SDM}, and ControlNet~\cite{controlnet}.
\subsection{Mask-to-Image examples}
\cref{fig:qual_combined} shows a comparison of synthetic images generated on the PanNuke~\cite{pannuke}, CoNIC~\cite{graham2024conic}, and MoNuSAC~\cite{verma2020multi} datasets. As illustrated in the figure, images generated by DiffMix appear very coarse with additional artifacts and fail to preserve the accurate stain colors observed in the original images. Consistent with observations reported by \cite{eccv_min}, we find that SDM-generated images display unrealistic color overlay artifacts.
% DiffMix struggles to preserve clear boundaries between cells, particularly in regions with high cell density, as highlighted by the red dotted square in \cref{fig:qual_combined}.
The color distribution of ControlNet-generated images appears highly inconsistent, being significantly inaccurate in some cases while better than others in certain instances. On the other hand, PathDiff accurately follows the cell mask and maintains the cell stain colors as seen in the original images. 
\label{sec: diverse_generation}

\subsection{Text-to-Image examples} 
\begin{figure*}[t]
    \centering
    \includegraphics[scale=0.77]{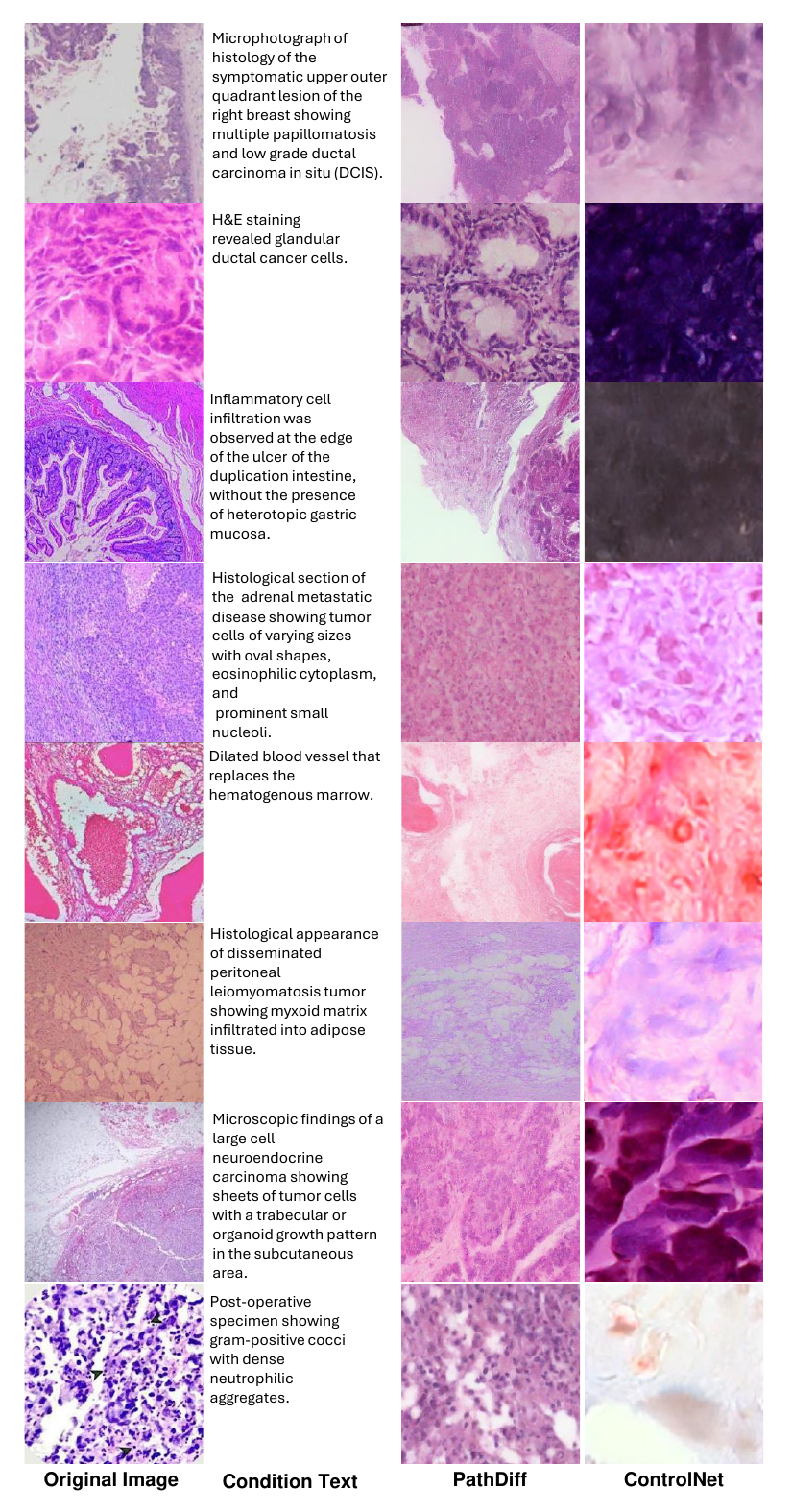}
    \caption{\textbf{Qualitative comparison} of synthetic images generated by PathDiff and ControlNet~\cite{controlnet} on the PathCap~\cite{PathCap} dataset.}
    \label{fig:qual_pathcap_supp}
\end{figure*}

\cref{fig:qual_pathcap_supp} shows samples generated by PathDiff and ControlNet. As with images conditioned on masks, ControlNet fails to preserve details in the original image and exhibits impractical colors uncommon in histopathology images. This explains the high FID and KID values compared to PathDiff in Tab.3 of the main paper.

\subsection{Unified Paired Conditions Sampling}
\begin{figure*}[t]
    \centering
    \includegraphics[scale=0.65]{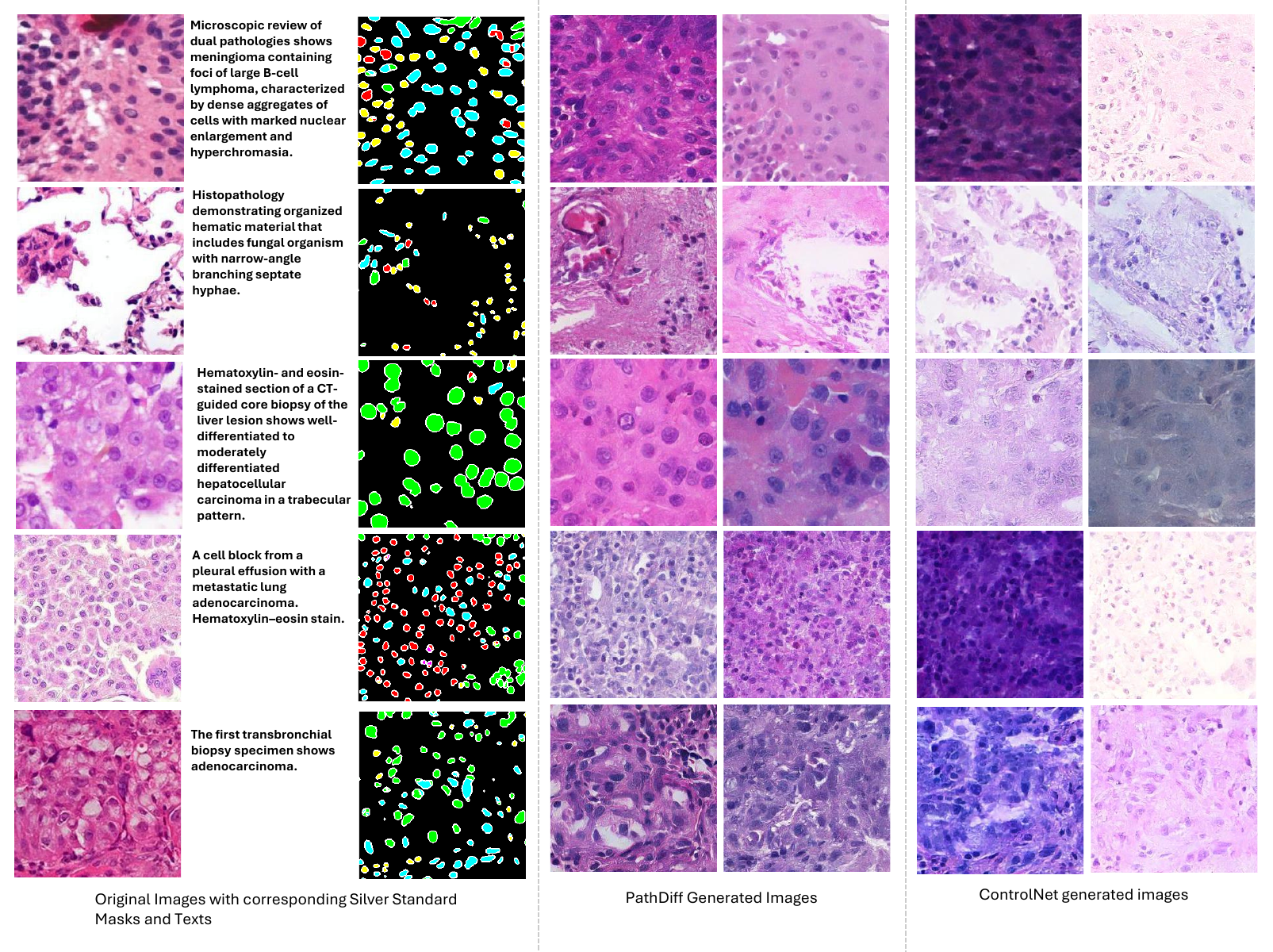}
    \caption{\textbf{Qualitative comparison of Pathdiff and ControlNet generated images on paired Silver standard masks and texts.}}
    \label{fig:silver_styd}
\end{figure*}

\cref{fig:silver_styd} shows images genertaed from paired Text and silver standard masks. PathDiff generated images incorporated guidance from both Text and Mask successfully and look significantly better than ControlNet.

\section{Domain Expert Assessment}
\label{sec:pathologists-experiment}

We acknowledge that traditional fidelity metrics like FID~\cite{FID} are only somewhat applicable to histological images as large image datasets like ImageNet~\cite{imagenet} would unlikely have images from this specific domain. Therefore, we conduct expert evaluation to validate the efficiency of the generated samples.
We surveyed two domain experts—a physician and a pathology researcher—to review the generated data and assess if the samples accurately reflect the characteristics of real specimens.

\begin{figure}[t]
        \centering
        \includegraphics[width=\linewidth]{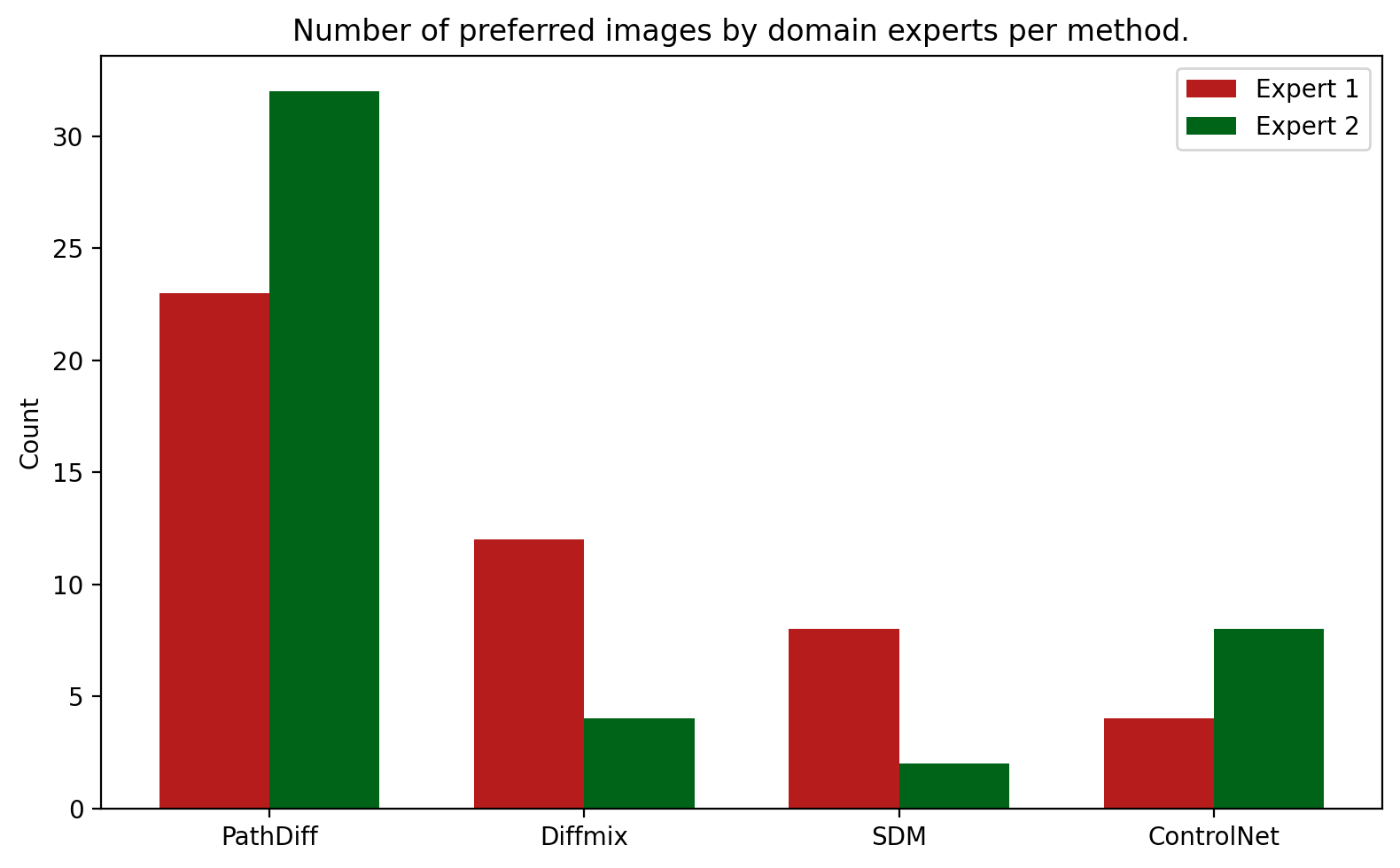}
        \caption{Both domain experts significantly preferred PathDiff generated images over other methods.}
        \label{fig:pref}
\end{figure}

\textbf{Image Preference Experiment:} 
We presented domain experts with a total of 200 synthetic images (Quadruplets of 50)  generated from
PathDiff, SDM~\cite{SDM}, ControlNet~\cite{controlnet}, DiffMix~\cite{Diffmix}. Each Quadruplet of images was generated using the same conditional mask. Domain experts were asked to choose one of the four images that looked most real. As shown in ~\cref{fig:pref}, both domain experts preferred PathDiff-generated images significantly more than the existing SOTA methods, indicating our generated images look more realistic to an expert eye compared to others.

\begin{table}[h!]
    \centering
    \begin{tabular}{lcc}
        \toprule
        & \multicolumn{2}{c}{\textbf{Predicted Label}} \\
        \cmidrule(lr){2-3}
        \textbf{Actual Label} & \textbf{Real} & \textbf{Synthetic} \\
        \midrule
        Synthetic & 15 & 12 \\
        Real      & 11 & 16 \\
        \bottomrule
    \end{tabular}
    \caption{Confusion matrix for a domain expert distinguishing Real vs.~Synthetic images.}
    \label{tab:user1_conf_matrix}
\end{table}

\textbf{Expert Turing Test:} In this experiment, a domain expert (physician) was presented with a total of 54 samples in equal numbers of real and synthetic images in random order. Real labels of images are hidden. We ask to choose whether the given image looks 

~\cref{tab:user1_conf_matrix}
shows domain expert's performance in distinguishing real from synthetic images. Out of 27 synthetic images, 12 were correctly identified, while 15 were mistaken as real. Among 27 real images, 11 were correctly labeled, with 16 falsely classified as synthetic. The overall accuracy was approximately 42.6. This indicates the user found it somewhat challenging to differentiate real from synthetic images.
\textit{real} or \textit{synthetic}. 

% As presented in the confusion matrix in \cref{fig:pathologist_1}~(b), domain experts struggled to consistently identify real and synthetic images, with many misclassifications occurring in both directions (real images labeled as synthetic and vice versa). Of all synthetic images presented, 31 were misclassified as real by domain experts, compared to only 21 correctly identified as synthetic. The high misclassification rate of synthetic images (approximately 60\% labeled as real) indicates that our synthetic images are realistic enough, making it difficult for domain experts to distinguish them from real images.

\section{Sampling Algorithm}
We use classifier-free guidance to sample from conditional and unconditional diffusion models to update the final score. \cref{alg:sampling} gives overview of the sampling. We either randomly pair the conditions from non-overlapping M2I and T2I datasets or generate silver standard masks for T2I dataset (or can generate caption/relevant text condition for M2I dataset).

\begin{algorithm}[t]
\footnotesize
\caption{\textbf{Unified Conditional Sampling}}
\label{alg:sampling}
\begin{algorithmic}[1]
\REQUIRE \textbf{$\mathbf{\omega}$}: guidance strength \\
         \textbf{Define} $\mathbf{c \in \left\{ (\varnothing_m, c_t), (c_m, \varnothing_t), (c_m, c_t) \right\}}$ \\
         \textbf{$\{1, \dots, T\}$}: timesteps with decreasing noise schedule $\alpha = \{\alpha_t\}_{t=1}^T$

\STATE \textbf{Initialize}: $\mathbf{z_T \sim \mathcal{N}(0, I)}$
\FOR{\textbf{$t = T$ to $1$}}
    \STATE \textcolor{gray}{$\triangleright$ Form classifier-free guided score at timestep $\mathbf{t}$} \\
    $\mathbf{\tilde{\epsilon}_\theta(z_t, t, c) = (1 + \omega) \epsilon_\theta(z_t, t, c) - \omega \epsilon_\theta(z_t, t)}$
    \STATE \textcolor{gray}{$\triangleright$ Denoise step to obtain intermediate sample $\mathbf{\tilde{z}_t}$} \\
    $\mathbf{\tilde{z}_t = \frac{z_t - \sqrt{1 - \bar{\alpha}_t} \, \tilde{\epsilon}_\theta(z_t, t, c)}{\sqrt{\bar{\alpha}_t}}}$
    \IF{\textbf{$t > 1$}}
        \STATE \textbf{Sample} $\mathbf{z_{t-1} \sim \mathcal{N}(\mu_\theta(z_t, \tilde{z}_t, t), \Sigma_\theta(z_t, t))}$
    \ELSE
        \STATE $\mathbf{z_0 = \tilde{z}_t}$
    \ENDIF
\ENDFOR
\STATE \textbf{return} $\mathbf{z_0}$
\end{algorithmic}
\end{algorithm}

\section{Considerations for \texorpdfstring{$p_{split}$}{p\_split}}
\label{p_split}
When training jointly on two datasets—Text-to-Image and Mask-to-Image—$p_{split}$ controls the proportion of data sampled from each of them. We evaluate performance with three values of $p_{split}$: 0.2, 0.5, and 0.8. Results using only text are shown in \Cref{tab:p_split_1}, mask-only conditioning in \cref{tab:p_split_2}, and both text and mask conditioning in \cref{tab:p_split_3}.

In these experiments, $p_{split}=0.5$ strikes a balance, explaining why we chose this value in the main paper. While it seems logical to assign a larger probability to the larger dataset to cover more of its samples, we found that $p_{split}=0.5$ works well in practice, ensuring samples from both datasets are included at least once per epoch.

\begin{table}[th!]
    \centering
    % \begin{minipage}{0.48\textwidth}
        \centering
        \resizebox{\linewidth}{!}{%
        \begin{tabular}{@{}lccccccc@{}}
            \toprule
            \multirow{2}{*}{\textbf{$p_{split}$}} & \multicolumn{3}{c}{\textbf{PathCap: Train}} & \multicolumn{3}{c}{\textbf{PathCap: Test}} \\ \cmidrule(lr){2-4} \cmidrule(lr){5-7}
                                             & \textbf{FID \(\boldsymbol{\downarrow}\)} & \textbf{KID \(\boldsymbol{\downarrow}\)} & \textbf{PLIP \(\boldsymbol{\uparrow}\)} & \textbf{FID \(\boldsymbol{\downarrow}\)} & \textbf{KID \(\boldsymbol{\downarrow}\)} & \textbf{PLIP \(\boldsymbol{\uparrow}\)} \\ \midrule
            $p_{split}=0.2$                 & 16.33 & 0.0624 & 24.43 & 15.74 & 0.0603 & 24.50 \\
            $p_{split}=0.5$                 & 18.52 & 0.0619 & 24.18 & 19.60 & 0.0644 & 24.05 \\
            $p_{split}=0.8$                 & 19.58 & 0.0649 & 24.34 & 18.87 & 0.0626 & 24.27 \\
            \bottomrule
        \end{tabular}}
        \caption{\textbf{Considerations for $p_{split}$.} CLIP-FID~\cite{clip, FID}, KID~\cite{kid}, and PLIP~\cite{PLIP} similarity scores for different $p_{split}$ values on PathCap~\cite{PathCap}, with text condition $c_t$ used for sampling. PLIP~\cite{PLIP} similarity scores on the real PathCap train and test splits are \textbf{26.34} and \textbf{26.56}, respectively, provided as a reference for comparison.}
        \label{tab:p_split_1}
    \end{table}
    % \end{minipage}%
    % \hfill
    % \begin{minipage}{0.48\textwidth}
    \begin{table}[th!]
        \centering
         % \setlength{\tabcolsep}{3pt} % Adjust column separation for compactness
        % {\fontsize{8}{12}\selectfont % Smaller font size for compactness
        \resizebox{0.75\linewidth}{!}{%
        \begin{tabular}{@{}lccccccc@{}}
            \toprule
            \multirow{2}{*}{\textbf{$p_{split}$}} & \multicolumn{2}{c}{\textbf{PanNuke: Train}} & \multicolumn{2}{c}{\textbf{PanNuke: Test}} \\ \cmidrule(lr){2-3} \cmidrule(lr){4-5}
                                             & \textbf{FID \(\boldsymbol{\downarrow}\)} & \textbf{KID \(\boldsymbol{\downarrow}\)}  & \textbf{FID \(\boldsymbol{\downarrow}\)} & \textbf{KID \(\boldsymbol{\downarrow}\)}  \\ \midrule
            $p_{split}=0.2$  & 7.36  & 0.0525 & 7.88 & 0.0559                                                                    \\
            $p_{split}=0.5$ & 6.94 & 0.0389 & 7.28 & 0.0415                                                                 \\
            $p_{split}=0.8$ & 8.57 & 0.0584 & 8.97 & 0.0707                                                                \\
            \bottomrule
        \end{tabular}}
        \caption{\textbf{Considerations for $p_{split}$.} CLIP-FID~\cite{clip, FID}, KID~\cite{kid} for different $p_{split}$ values on PanNuke~\cite{pannuke} dataset. Only mask condition $c_m$ was used for sampling.}
        \label{tab:p_split_2}
    % \end{minipage}
\end{table}

\begin{table*}[th!]
    \centering
    % \setlength{\tabcolsep}{3pt} % Adjust column separation for compactness
    % {\fontsize{8}{12}\selectfont % Smaller font size for compactness
    \resizebox{0.9\linewidth}{!}{%
    \begin{tabular}{@{}lccccccccccccccc@{}}
        \toprule
        \multirow{3}{*}{\textbf{$p_{split}$}} & \multicolumn{4}{c}{\textbf{PanNuke}} & \multicolumn{6}{c}{\textbf{PathCap}} \\ \cmidrule(lr){2-5} \cmidrule(lr){6-11}
                                         & \multicolumn{2}{c}{\textbf{Train}} & \multicolumn{2}{c}{\textbf{Test}} & \multicolumn{3}{c}{\textbf{Train}} & \multicolumn{3}{c}{\textbf{Test}} \\ \cmidrule(lr){2-3} \cmidrule(lr){4-5} \cmidrule(lr){6-8} \cmidrule(lr){9-11}
                                         & \textbf{FID \(\boldsymbol{\downarrow}\)} & \textbf{KID \(\boldsymbol{\downarrow}\)} & \textbf{FID \(\boldsymbol{\downarrow}\)} & \textbf{KID \(\boldsymbol{\downarrow}\)} & \textbf{FID \(\boldsymbol{\downarrow}\)} & \textbf{KID \(\boldsymbol{\downarrow}\)} & \textbf{PLIP \(\boldsymbol{\uparrow}\)} & \textbf{FID \(\boldsymbol{\downarrow}\)} & \textbf{KID \(\boldsymbol{\downarrow}\)} & \textbf{PLIP \(\boldsymbol{\uparrow}\)} \\ \midrule
        $p_{split}=0.2$                          & 10.25        & 0.0672        &  11.99       &  0.0800       &  15.21       &  0.0846      & 23.01 & 16.07    & 0.0956       &  22.81 \\
        $p_{split}=0.5$                          & 11.03       & 0.0718        &    12.32    &   0.0952     &  14.39      &  0.0884      &  23.01 &  14.26     &  0.1059       & 22.80 \\
        $p_{split}=0.8$                          &  9.723       &   0.0729      & 10.37 & 0.0862        & 12.78 & 0.0955 & 22.97 & 12.53  & 0.1107 & 22.70 \\
        \bottomrule
    \end{tabular}}
    \caption{\textbf{Consideration for $p_{split}$.} CLIP-FID~\cite{clip, FID}, KID~\cite{FID}, and PLIP~\cite{PLIP} similarity scores for different $p_{split}$ values for PanNuke~\cite{pannuke} and PathCap~\cite{PathCap}. We used both text $c_t$ and mask $c_m$ for sampling.}
    \label{tab:p_split_3}
\end{table*}

\section{In-Domain FID Results}
For a more faithful assessment of pathology image quality, we compute an in-domain FID using the CONCH~\cite{conch_nature} encoder rather than relying solely on CLIP or Inception-based features, which were trained on general natural images and may not capture the nuances of histopathology. CONCH~\cite{conch_nature} is a foundation model trained on large pathology image-text pairs.

\begin{table}[ht]
  \centering
  \begin{tabular}{@{}lccc@{}}
    \toprule
    \textbf{Method} & \textbf{PanNuke} & \textbf{CoNIC} & \textbf{MoNuSAC} \\
    \midrule
    Diffmix                         & 119.35                   & 257.92                   & 290.27                    \\
    SDM                             & 177.30                   & \underline{143.48}                   & \underline{166.65}        \\
    ControlNet                      & \underline{121.57}                   & 174.74                   & 277.96                    \\
    % PathDiff ($p_{\mathrm{split}}=0$)   & \underline{82.41}         & \underline{110.82}        & 308.95                    \\
    PathDiff & \textbf{53.74}            & \textbf{91.21}            & \textbf{121.61}           \\
    \bottomrule
  \end{tabular}
  \caption{\textbf{Comparison of CONCH-FID} across training splits for PanNuke~\cite{pannuke}, CoNIC~\cite{graham2024conic}, and MoNuSAC~\cite{verma2020multi}. PathDiff is trained jointly with T2I dataset: PathCap~\cite{PathCap}. ControlNet~\cite{controlnet} uses SD~\cite{stablediffusion} backbone trained on the PathCap dataset.}
  \label{tab:m2i_fid_split0}
\end{table}

\begin{table}[ht]
  \centering
  \begin{tabular}{@{}lccc@{}}
    \toprule
    \textbf{Method} & \textbf{PanNuke} & \textbf{CoNIC} & \textbf{MoNuSAC} \\
    \midrule
    ControlNet                      & 347.77                   & 343.39                   & 331.61                    \\
    % PathDiff ($p_{\mathrm{split}}=1$)   & \textbf{142.89}           & \textbf{142.89}           & \underline{142.89}        \\
    PathDiff & \textbf{153.77}        & \textbf{156.21}        & \textbf{141.88}           \\
    \bottomrule
  \end{tabular}
  \caption{\textbf{Comparison of CONCH-FID} on training splits for T2I dataset: PathCap. PathDiff is jointly trained with three M2I datasets: PanNuke~\cite{pannuke}, CoNIC~\cite{graham2024conic}, and MoNuSAC~\cite{verma2020multi}.}
  \label{tab:m2i_fid_split1}
\end{table}

\section{Mask Depth Ablation}
We test the effect of using two types of conditioning mask, first cell type mask and other is instance mask. We generate mask edges from instance mask using image processing technics. Using both masks generates better quality images as seen in~\cref{tab:temporal_abla}, subsequently we use the mask depth of 6. We simply increase the channel size of the Mask embedder $M$ and concatenate two masks as input. 
\begin{table}[ht]
  \centering
  \begin{tabular}{@{}lcccc@{}}
    \toprule
    \textbf{\# mask\_depth} 
      & \multicolumn{4}{c}{\textbf{PanNuke Test}} \\
    \cmidrule(lr){2-5}
      & \textbf{IP \(\uparrow\)} & \textbf{IR} \(\uparrow\) & \textbf{CONCH FID} \(\downarrow\) & \textbf{KID} \(\downarrow\) \\
    \midrule
    \textbf{3} & 0.79 & 0.47 & 102.37 & 0.0644 \\
    \textbf{6} & 0.72 & 0.77 &   7.21 & 0.0415 \\
    \bottomrule
  \end{tabular}
    \caption{\textbf{Mask depth ablation} on PanNuke test split}
    \label{tab:temporal_abla}
\end{table}

\section{Choice of Pretrained Checkpoints}
We tried different pretrained checkpoints choices in three module components: VAE, Text-Encoder,and Unet.

\paragraph{Finetuning VAE}
The reconstruction performance of VAEs~\cite{Pathldm,stablediffusion} plays a crucial role in the fidelity of generated images. Losses introduced during the compression and decompression stages in VAEs compound with the denoising process losses in subsequent stages, directly impacting the quality of the generated images.

Initially, we used the VQ-VAE from~\cite{Pathldm}, which was trained on the TCGA-BRCA~\cite{tcga_brca} dataset containing whole-slide images (WSIs) exclusively from breast tissues. While this VAE outperforms the one from~\cite{stablediffusion}, which was trained on natural images, its applicability is limited as it lacks representation of diverse tissue types. We fine-tuned the VAE on the datasets used in this work, including PanNuke~\cite{pannuke}, PathCap~\cite{PathCap}, CoNIC~\cite{graham2024conic}, and MoNuSAC~\cite{verma2020multi}.

As demonstrated in ~\cref{tab:vae_finetune}, fine-tuning the VAE on these datasets results in improvements across all reconstruction and generation metrics. However, these improvements, while consistent, are relatively modest.

\begin{table}[th!]
    \centering
    \resizebox{\linewidth}{!}{%
    \begin{tabular}{@{}lcccc@{}}
        \toprule
        \multirow{2}{*}{VAE Trained on} & \multicolumn{4}{c}{\textbf{Metrics}} \\ \cmidrule(lr){2-5}
                                             & \textbf{LPIPS \(\boldsymbol{\downarrow}\)} & \textbf{SSIM \(\boldsymbol{\uparrow}\)} & \textbf{MSE \(\boldsymbol{\downarrow}\)} & \textbf{FID \(\boldsymbol{\downarrow}\)} \\ \midrule
        \textbf{TCGA-BRCA \cite{tcga_brca}}                         &        0.0462        &        0.7962        &    0.0084          &     6.94         \\ 
        \textbf{Datasets: D}                         &     0.0429            &   0.8212            &     0.0070         &  6.31            \\ 
        \bottomrule
    \end{tabular}}
    \caption{\textbf{Effect of fine-tuning VAE on datasets D:} PanNuke~\cite{pannuke}, PathCap~\cite{PathCap}, MoNuSAC~\cite{verma2020multi}, and CoNIC~\cite{graham2024conic}.}
    \label{tab:vae_finetune}
\end{table}

\paragraph{Text Encoder}
To evaluate text–image alignment on the PathCap training set, we compared the similarity between each image and its corresponding report using two embedding methods. The CLIP-based similarity score was 21.56, while the PLIP-based score reached 26.30; clearly demonstrating that PLIP embeddings achieve stronger alignment between images and text. Therefore we used PLIP text encoder checkpoint in our experiments.

\paragraph{Denoising U-net}
We fine-tune the U-Net on the PathCap text-to-image dataset starting from the TCGA-BRCA checkpoint provided by \cite{graikos2024learned}.
The results are summarized in~\cref{tab:unet_ckpt_comparison}. We see improved CLIP-FID and KID scores for mask-to-image generation on PanNuke, as well as a higher PLIP similarity score for text-to-image generation on PathCap.

\begin{table}[ht]
  \centering
  \begin{tabular}{@{}lccc@{}}
    \toprule
    \textbf{U-Net Checkpoint} & \textbf{CLIP FID ↓} & \textbf{PLIP Score ↑} & \textbf{KID ↓} \\
    \midrule
    \textbf{TCGA-BRCA} & 14.44 & 21.79 & 0.1284 \\
    \textbf{PathCap}   &  7.21 & 24.05 & 0.0410 \\
    \bottomrule
  \end{tabular}
    \caption{\textbf{Comparison of U-Net checkpoints} on CLIP FID, PLIP score, and KID}
  \label{tab:unet_ckpt_comparison}
\end{table}

\section{Significance Test on Downstream Task:}
\begin{figure*}[t]
        \centering        \includegraphics[width=\linewidth]{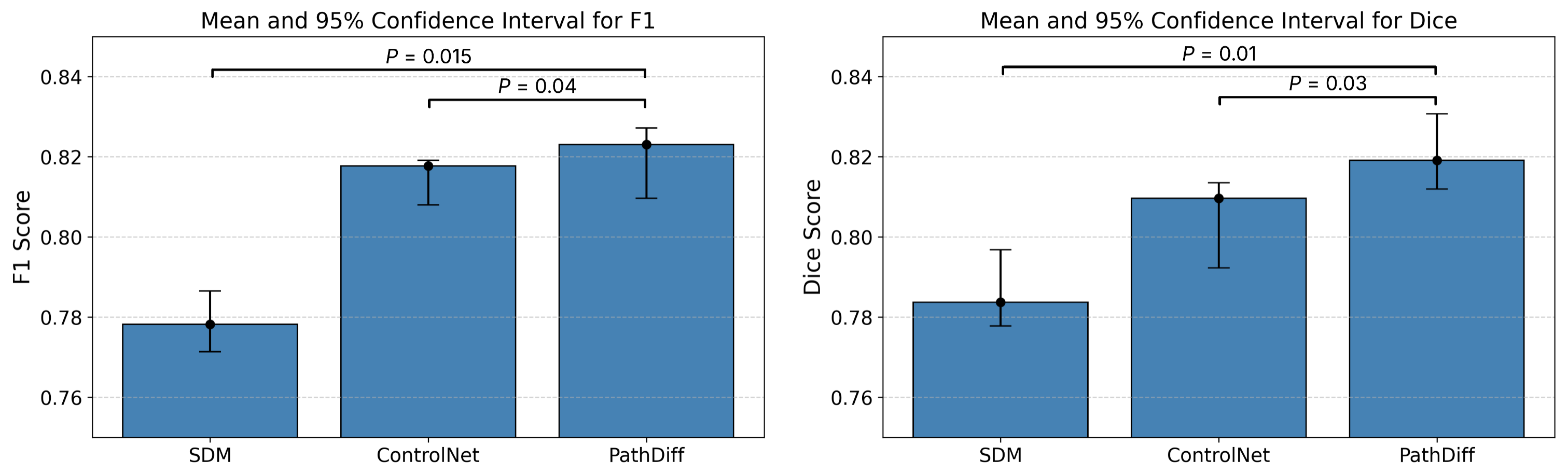}
        \caption{\textbf{Pairwise significance test}, $p<0.05$ indicates that PathDiff augmentation set helps improve downstream classification and segmentation tasks statistically significantly as compared to other methods.}
        \label{fig:statsical_test}
\end{figure*}

We validate that PathDiff’s higher downstream scores aren’t due to chance by running a paired permutation test. We randomly swap method labels within each test example and compare mean F1/Dice across 1000 trials. The resulting p-values are $<0.05$ confirming PathDiff’s gains are statistically significant as seen in \cref{fig:statsical_test}.

\section{Generation Performance on Hard Pathology Cases}
\label{sec:hard-pathology-cases}

To assess how well PathDiff handles challenging, clinically significant images, we split our test set into “pathological” (reports mentioning “carcinoma”) and “non-pathological” cases (reports describing benign findings). Table~\ref{tab:hard-pathology} compares FID, KID, and PLIP scores for each group. Although overall image fidelity remains similar, lower PLIP score for pathological cases suggest that images with malignant features are marginally more difficult to synthesize than benign ones.

\begin{table}[ht]
  \centering
  \caption{Performance on pathological vs.\ non-pathological cases}
  \label{tab:hard-pathology}
  \begin{tabular}{@{}lccc@{}}
    \toprule
    \textbf{Case Type}        & \textbf{FID ↓} & \textbf{KID ↓} & \textbf{PLIP ↑} \\
    \midrule
    Pathological              & 19.97          & 0.04           & 23.15          \\
    Non-pathological          & 20.10          & 0.06           & 24.13          \\
    \bottomrule
  \end{tabular}
\end{table}

\section{Details on training previous works:}
Diffmix and SDM are trained on M2I datasets only. We use their official repositories to refer to their code. For both DiffMix and SDM we use same training settings for all M2I datasets that of PanNuke~\cite{pannuke} in ~\cite{Diffmix}. For ControlNet we pre-trained SD\cite{stablediffusion} model
on T2I data first and then used only M2I data for finetuning, as recommended in the official controlnet tutorial.

\section{Computational Costs:}
Since PathDiff only trains Unet encoder and shallow mask embedder, training costs remain modest, even for joint training. PathDiff trains 694 M parameters. Training time for the
largest dataset combination (PathCap~\cite{PathCap} + PanNuke~\cite{pannuke}) is 30 Hours on 4 NVIDIA A6000 GPUs. Sampling ~6,300(train split of PanNuke) images takes 3.5-4.5 hours.

\section{Survey Tool:}
\begin{figure*}[t]
        \centering        \includegraphics[width=\linewidth]{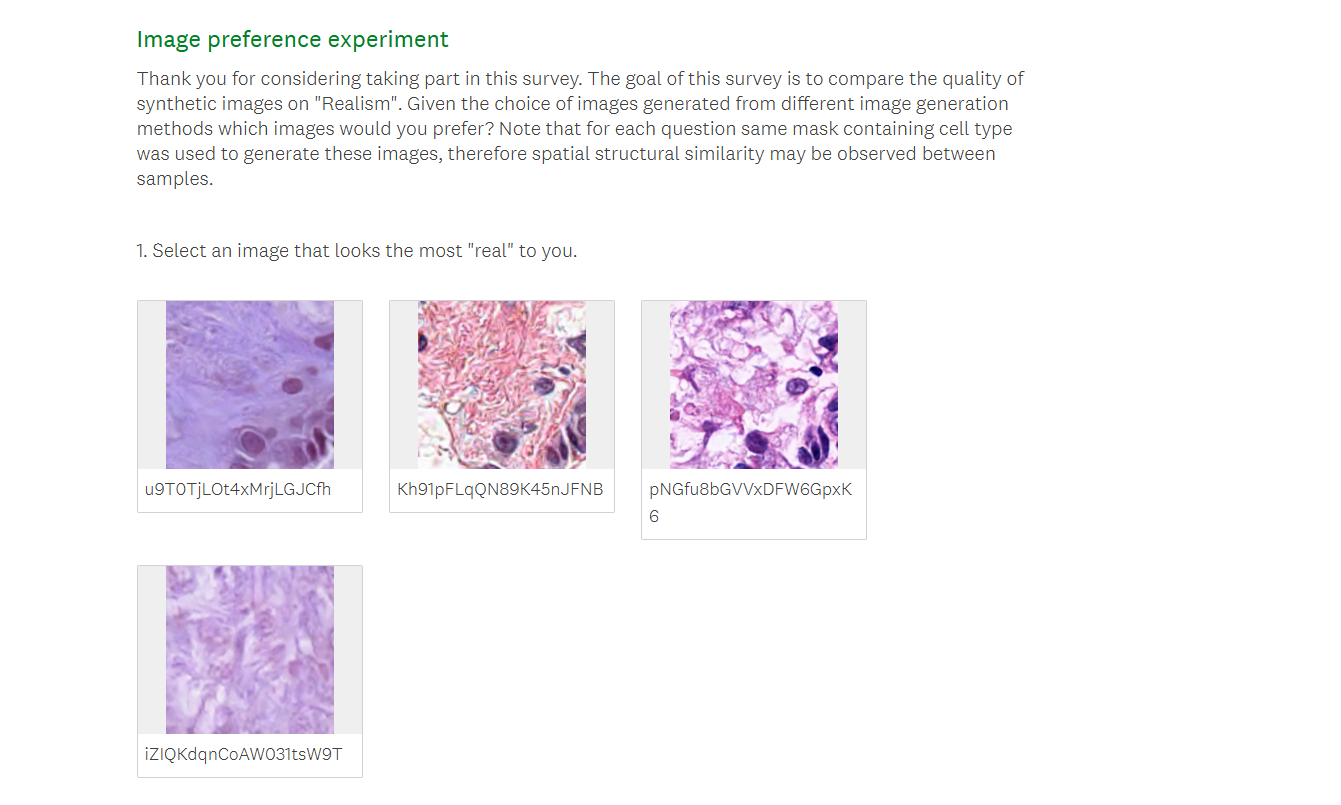}
        \caption{Interactive web interface used for domain expert image preference experiment.}
        \label{fig:img_pref_survey}
\end{figure*}

\begin{figure*}[t]
        \centering        \includegraphics[width=\linewidth]{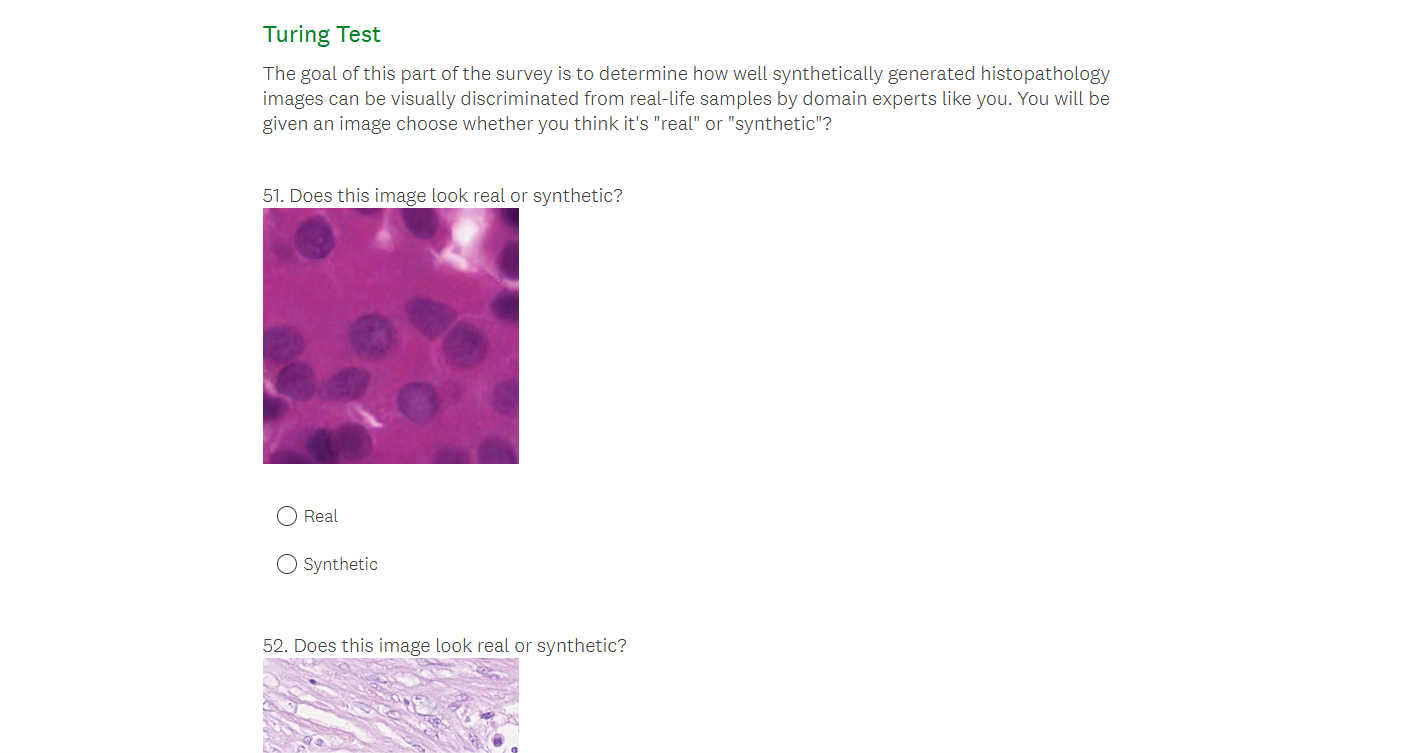}
        \caption{Interactive web interface used for domain expert Turing test.}
        \label{fig:img_turing_test_survey}
\end{figure*}

We used an interactive web-based tool to conduct a domain experts survey. Clear instructions were given to evaluate the images. \cref{fig:img_pref_survey} and \cref{fig:img_turing_test_survey} show the web interface used for the domain expert image preference experiment and the Turing test respectively.

\end{document}